%% file: main.tex
\documentclass[10pt,twocolumn,letterpaper]{article}

\usepackage{cvpr}              

\def\cvprPaperID{11854} 
\def\confName{CVPR}
\def\confYear{2023}

\usepackage[utf8]{inputenc} 
\usepackage[T1]{fontenc}    
\usepackage{hyperref}       
\usepackage{url}            
\usepackage{booktabs}       
\usepackage{amsfonts}       
\usepackage{nicefrac}       
\usepackage{microtype}      
\usepackage[dvipsnames]{xcolor}         
\usepackage[accsupp]{axessibility}  

\usepackage{graphicx}
\usepackage{amsmath}
\usepackage{amssymb}
\usepackage{multirow}
\usepackage[ruled, lined]{algorithm2e}

\usepackage{hyperref}
\usepackage{array}

\usepackage[accsupp]{axessibility}

\newcommand{\nameemoji}{\includegraphics[height=.8\baselineskip]{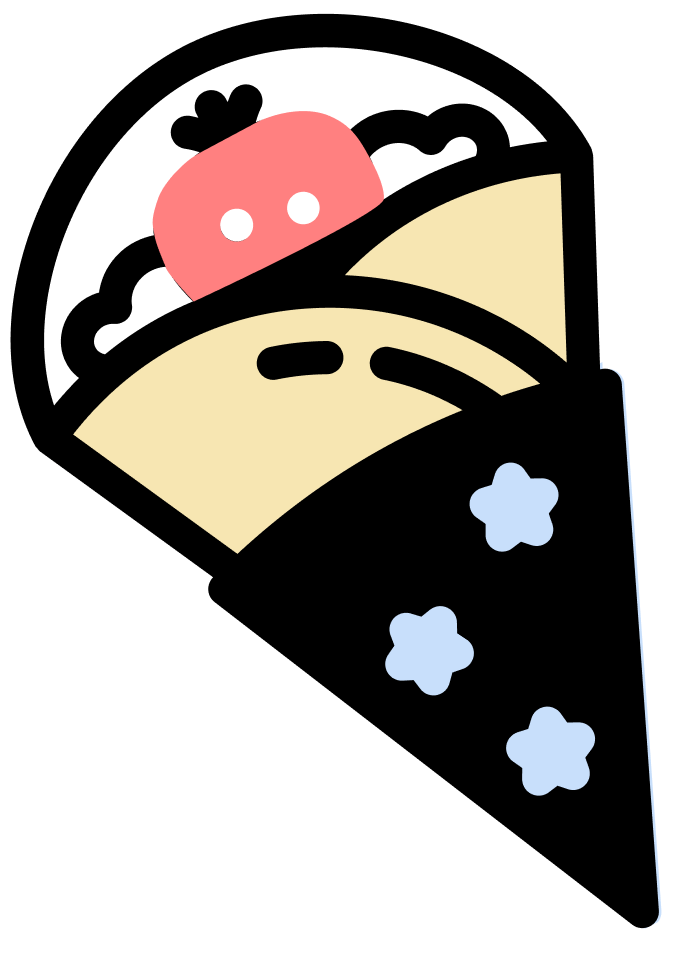} CREPE}
\newcommand{\name}{CREPE}
\newcommand{\D}{\mathcal{D}}
\newcommand{\hnerror}[1]{\textcolor{red}{#1}}
\newcommand{\correct}[1]{{\textcolor{ForestGreen}{#1}}}

\title{
\nameemoji: Can Vision-Language Foundation Models Reason Compositionally?
}

\author{Zixian Ma$^1$*, Jerry Hong$^1$*, Mustafa Omer Gul$^2$*, Mona Gandhi$^3$, Irena Gao$^1$, Ranjay Krishna$^4$\\
Stanford University$^1$, Cornell University$^2$, University of Pennsylvania$^3$, University of Washington$^4$\\
{\tt\small \{zixianma, jerryhong, irena\}@cs.stanford.edu mog29@cornell.edu mona09@seas.upenn.edu } \\ {\tt\small ranjay@cs.washington.edu }
}

\begin{document}

\maketitle

\def\thefootnote{*}\footnotetext{Equal contribution}\def\thefootnote{\arabic{footnote}}

\begin{abstract}

\input{sections/00_abstract}
\end{abstract}

\input{sections/01_introduction}
\input{sections/02_related_work} 
\input{sections/03_compositionality}

\input{sections/04_methods}
\input{sections/05_experiments}
\input{sections/06_discussion}
\input{sections/07_conclusion}

\newpage
{\small
\bibliographystyle{ieee_fullname}
\bibliography{egbib}
}

\newpage
\appendix
\input{sections/08_appendix.tex}

\end{document}

%% file: sections/00_abstract.tex
A fundamental characteristic common to both human vision and natural language is their compositional nature. Yet, despite the performance gains contributed by large vision and language pretraining, we find that—across 7 architectures trained with 4 algorithms on massive datasets—they struggle at compositionality.
To arrive at this conclusion, we introduce a new compositionality evaluation benchmark, \nameemoji{}, which measures two important aspects of compositionality identified by cognitive science literature: systematicity and productivity. To measure systematicity, \name{} consists of a test dataset containing over $370K$ image-text pairs and three different seen-unseen splits. The three splits are designed to test models trained on three popular training datasets: CC-12M, YFCC-15M, and LAION-400M. We also generate $325K$, $316K$, and $309K$ hard negative captions for a subset of the pairs. To test productivity, \name{} contains $17K$ image-text pairs with nine different complexities plus $183K$ hard negative captions with atomic, swapping and negation foils. The datasets are generated by repurposing the Visual Genome scene graphs and region descriptions and applying handcrafted templates and GPT-3.
For systematicity, we find that model performance decreases consistently when novel compositions dominate the retrieval set, with Recall@1 dropping by up to $12\%$. For productivity, models' retrieval success decays as complexity increases, frequently nearing random chance at high complexity. These results hold regardless of model and training dataset size. 

%% file: sections/01_introduction.tex
\input{figures/pull_figure}

\section{Introduction}
Compositionality, the understanding that ``the meaning of the whole is a function of the meanings of its parts''~\cite{cresswell1973logics}, is held to be a key characteristic of human intelligence. In language, the whole is a sentence, made up of words. In vision, the whole is a scene, made up of parts like objects, their attributes, and their relationships~\cite{krishnavisualgenome,ji2020action}.
Through compositional reasoning, humans can understand new scenes and generate complex sentences by combining known parts~\cite{janssen1997compositionality,hupkes2020compositionality,bottou2014machine}.
Despite compositionality's importance, 
there are no large-scale benchmarks directly evaluating whether vision-language models can reason compositionally.
These models are pretrained using large-scale image-caption datasets~\cite{thomee2016yfcc100m,schuhmann2021laion,sharma2018conceptual}, and are already widely applied for tasks that benefit from compositional reasoning, including retrieval, text-to-image generation, and open-vocabulary classification~\cite{conwell2022testing,ramesh2022hierarchical,saharia2022photorealistic}.
Especially as such models become ubiquitous ``foundations'' for other models~\cite{bommasani2021opportunities}, it is critical to understand their compositional abilities.

\input{figures/systematicity_figure}
Previous work has evaluated these models using image-text retrieval~\cite{jia2021scaling,radford2021learning,Zhang_2021_CVPR}. However, the retrieval datasets used either do not provide controlled sets of negatives~\cite{lin2014microsoft,thomee2016yfcc100m} or study narrow negatives which vary along a single axis (\eg permuted word orders or single word substitutions as negative captions)~\cite{parcalabescu-etal-2022-valse,thrush2022winoground,shekhar-etal-2017-foil,https://doi.org/10.48550/arxiv.2106.09141}.
Further, these analyses have also not studied how retrieval performance varies when generalizing to unseen compositional combinations, or to combinations of increased complexity.

We introduce \nameemoji{} (Compositional REPresentation Evaluation): a new large-scale benchmark to evaluate two aspects of compositionality: \textit{systematicity} and \textit{productivity} (Figure~\ref{fig:pull_figure}). Systematicity measures how well a model is able to represent seen versus unseen atoms and their compositions. Productivity studies how well a model can comprehend an unbounded set of increasingly complex expressions. \name{} uses Visual Genome's scene graph representation as the compositionality language~\cite{krishnavisualgenome} and constructs evaluation datasets using its annotations.
To test systematicity, we parse the captions in three popular training datasets, CC-12M~\cite{changpinyo2021conceptual}, YFCC-15M~\cite{thomee2016yfcc100m}, and LAION-400M~\cite{schuhmann2021laion}, to identify atoms (objects, relations, or attributes) and compounds (combinations of atoms) present in each dataset.
For each training set, we curate corresponding test sets containing $385K$, $385K$ and $373K$ image-text pairs respectively, 
with splits checking generalization to seen compounds, unseen compounds, and unseen atoms.
To test productivity, \name{} contains $17K$ image-text pairs split across nine levels of complexity, as defined by the number of atoms present in the text.
Examples across all datasets are paired with various hard negative types to ensure the legitimacy of our conclusions.

Our experiments—across 7 architectures trained with 4 training algorithms on massive datasets—find that vision-language models struggle at compositionality, with both systematicity and productivity. 
We present six key findings: first, our systematicity experiments find that models' performance consistently drops between seen and unseen compositions; second, we observe larger drops for models trained on LAION-400M (up to a $12\%$ decrease in Recall@1); third, our productivity experiments indicate that retrieval performance degrades with increased caption complexity; fourth, we find no clear trend relating training dataset size to models' compositional reasoning; fifth, model size also has no impact; finally, models' zero-shot ImageNet classification accuracy correlates only with their absolute retrieval performance on the systematicity dataset but not systematic generalization to unseen compounds or to productivity.
\footnote{We release our datasets, and code to generate and evaluate on our test sets at \url{https://github.com/RAIVNLab/CREPE}.}

%% file: figures/pull_figure.tex
\begin{figure}[h!]
     \centering
     \includegraphics[width=\linewidth]{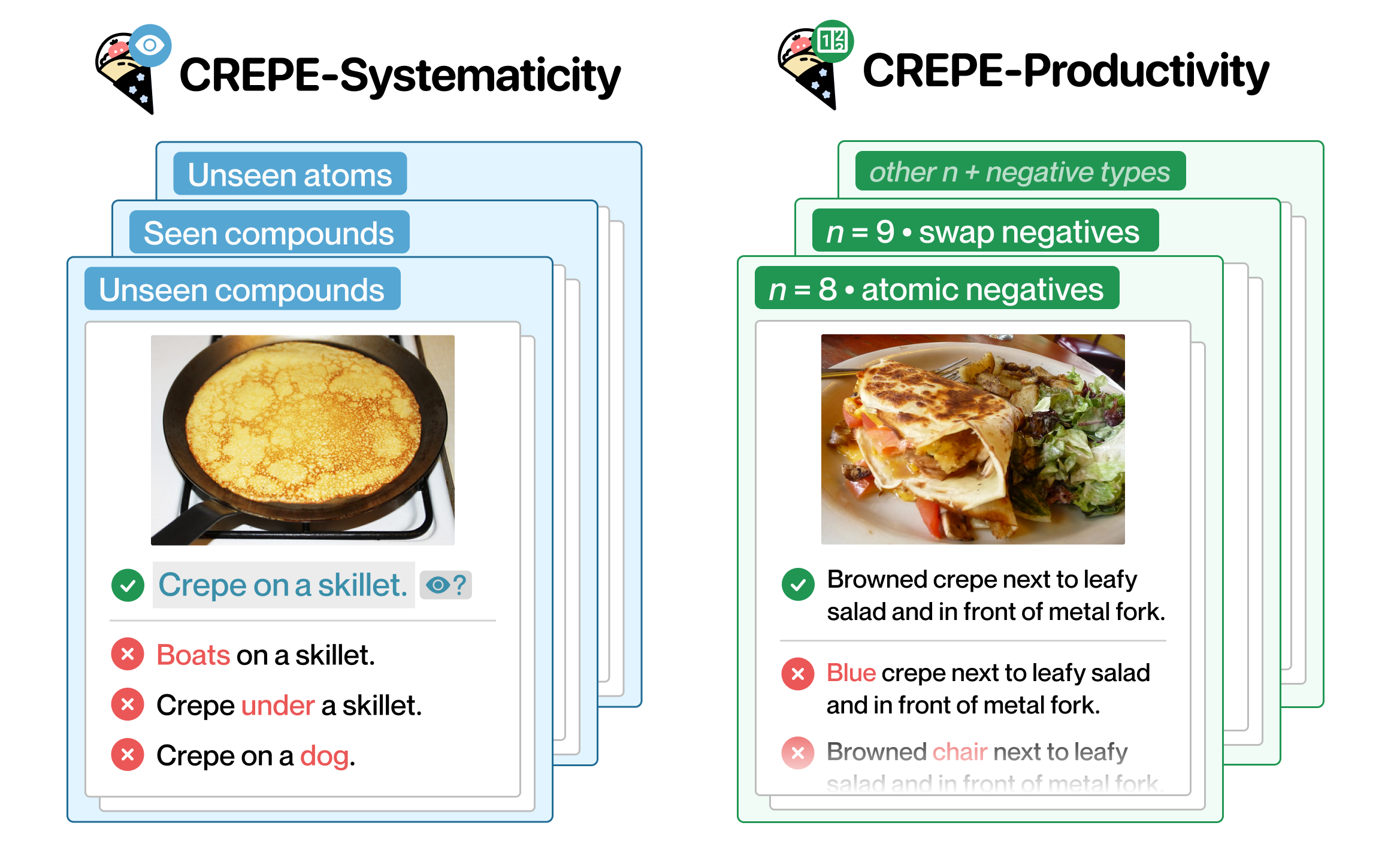}
     \caption{We introduce \nameemoji{}, a benchmark to evaluate whether vision-language foundation models demonstrate two fundamental aspects of compositionality: systematicity and productivity. To evaluate systematicity, \name{} utilizes Visual Genome and introduces three new test datasets for the three popular pretraining datasets: CC-12M, YFCC-15M, and LAION-400M. These enable evaluating models' abilities to systematically generalize their understanding to seen compounds, unseen compounds, and even unseen atoms. To evaluate productivity, \name{} introduces examples of nine complexities, with three types of hard negatives for each.
     }
     \label{fig:pull_figure}
 \end{figure}

%% file: figures/systematicity_figure.tex
\begin{figure*}[t]
     \centering
     \includegraphics[width=\linewidth]{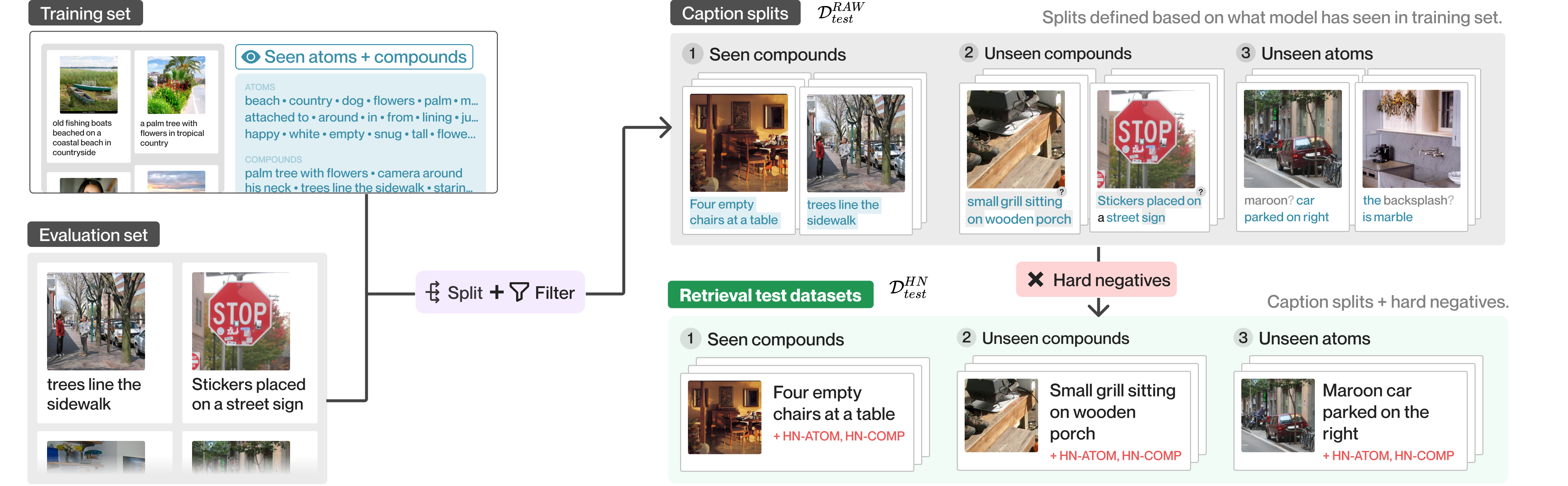}
     \caption{An overview of the \textbf{systematicity} retrieval set generation process. First, a model's image-caption training set is parsed to identify what atoms and compounds the model has seen. Then, an evaluation set is divided into three compositional splits according to whether the model has seen all the compounds (Seen Compounds), only all the atoms of the caption (Unseen Compounds), or neither (Unseen Atoms). Finally, hard negative captions $\textsc{HN-Atom}$ and $\textsc{HN-Comp}$ are generated for the hard negatives retrieval set $\D_{test}^{HN}$ .}
     \label{fig:systematicity}
 \end{figure*}

%% file: sections/02_related_work.tex
\label{sec:rw}
\section{Related Work}
Our work lies within the field of evaluating foundation models. Specifically, we measure visio-linguistic compositionality. To do so, we create a retrieval benchmark with hard negatives.

\noindent\textbf{Contrastive Image-Text Pretraining.} 
The recently released contrastively trained CLIP model~\cite{radford2021learning} has catalyzed a wide array of work at the intersection of Computer Vision and Natural Language Processing. Since its release, CLIP has enabled several tasks, ranging from semantic segmentation to image captioning, many of which have remarkable zero-shot capability~\cite{radford2021learning,li2022languagedriven,tewel2021zero, subramanian2022reclip,https://doi.org/10.48550/arxiv.2109.02748,pmlr-v168-cui22a}. CLIP has been used as a loss function within image synthesis applications~\cite{https://doi.org/10.48550/arxiv.2112.05744, Patashnik_2021_ICCV, xu2022ppe, Li_2022_CVPR, zhou2021lafite, Jain_2021_ICCV}, acted as an automated evaluation metric~\cite{park2021benchmark, hessel-etal-2021-clipscore}, used successfully as a feature extractor for various vision and language tasks~\cite{shen2022how}, and incorporated into architectures for various tasks including dense prediction and video summarization~\cite{9880269,rao2021denseclip,cosnet2022cvpr,shridhar2021cliport,https://doi.org/10.48550/arxiv.2107.00650,https://doi.org/10.48550/arxiv.2201.06696}. This success has also encouraged the design of other contrastive vision and language pretraining algorithms for image~\cite{li2021align,https://doi.org/10.48550/arxiv.2205.14459,https://doi.org/10.48550/arxiv.2204.07441,https://doi.org/10.48550/arxiv.2203.00048,li-etal-2022-unimo,yu2022coca,yang2022vision,https://doi.org/10.48550/arxiv.2112.04482,li2022blip} and video domains~\cite{9878891,https://doi.org/10.48550/arxiv.2109.08472,xu-etal-2021-videoclip}. Our work evaluates how well such contrastively trained models capture a fundamental property present in human vision and language: compositionality.

\noindent\textbf{Compositionality.}
Compositionality allows us to comprehend an infinite number of scenes and utterances~\cite{lake_ullman_tenenbaum_gershman_2017}. For an AI model, compositionality would not only allow for systematic, combinatorial generalization, but would also confer benefits such as controllability~\cite{bommasani2021opportunities}. This promise prompted a wealth of work on both designing~\cite{higgins2017scan,https://doi.org/10.48550/arxiv.1511.02799,https://doi.org/10.48550/arxiv.1803.03067} and evaluating ~\cite{hudson2018gqa,GrundeMcLaughlin2021AGQA,lake2018generalization,suhr-etal-2019-corpus,gandhi2022measuring} compositional models. 
In our work, we focus on two aspects of compositionality: systematicity and productivity.
While there is a plethora of benchmarks for systematic generalization within Computer Vision~\cite{https://doi.org/10.48550/arxiv.1612.06890,bahdanau2019closure,bogin-etal-2021-covr,GrundeMcLaughlin2021AGQA} and Machine Learning~\cite{lake2018generalization,10.5555/3495724.3497391,keysers2020measuring}, the subject has been almost unexplored for vision-language models, largely due to lack of benchmarks complementary to the different large-scale training datasets. To address this, \name{} provides a benchmark with three different datasets to evaluate the compositional generalization of vision-language models.
Productivity, on the other hand, has been studied only for specialized tasks~\cite{GrundeMcLaughlin2021AGQA} or toy domains~\cite{lake2018generalization,10.5555/3495724.3497391, hupkes2020compositionality}. \name{} evaluates productivity by using an image-text retrieval task featuring captions of varying compositional complexity.

\noindent\textbf{Evaluation with hard negatives.}
Like us, past work evaluating models has commonly designed tasks featuring hard negatives to isolate particular model capabilities while overcoming the limitations of prior evaluation tasks. 
Using atomic foils that replace an atom in the image or text with a distractor has been the most common strategy~\cite{shekhar-etal-2017-foil,https://doi.org/10.48550/arxiv.1901.06595,https://doi.org/10.48550/arxiv.2106.09141,https://doi.org/10.48550/arxiv.2003.00403,bogin-etal-2021-covr, park-etal-2022-exposing, parcalabescu-etal-2022-valse}.
Notably, Park \textit{et al.}~\cite{park-etal-2022-exposing} targets verbs and person entities in videos;
COVR~\cite{bogin-etal-2021-covr} studies question answering with distractor images; VALSE~\cite{parcalabescu-etal-2022-valse} targets linguistic phenomena such as existence, cardinality and the recognition of actions and spatial relationships.
Another strategy has been to swap atoms within a caption to test whether models behave akin to a bag-of-words~\cite{akula-etal-2020-words,parcalabescu-etal-2022-valse,thrush2022winoground}. In particular, Winoground~\cite{thrush2022winoground} introduces a set of $800$ human edited negatives to evaluate compositionality; it is the closest related work to us. We complement Winoground by scaling it up by three orders of magnitude, by decomposing compositionality into systematicity and productivity, and by studying a variety of different types of hard negatives.

%% file: sections/03_compositionality.tex
\input{figures/productivity_figure}

\section{Compositional evaluation}


The following section builds from the formally vacuous principle of compositionality to a well-defined evaluation scheme~\cite{hupkes2020compositionality}. First, we establish the syntax and semantics of the composed language (Section~\ref{sec:language}). Then, we define expected behaviors from a model that achieves comprehension of said language (~\ref{sec:systematicity},~\ref{sec:prodictivity}). Finally, we establish how to empirically measure those behaviors via retrieval (~\ref{sec:retrieval}). 

\subsection{Compositional language of visual concepts}
\label{sec:language}
To evaluate vision-language models, we find that a compositional language consisting of \textit{scene graph} visual concepts is an appropriate foundation~\cite{krishnavisualgenome}. Accordingly, an \textit{atom} $A$ is defined as a singular visual concept, corresponding to a single scene graph node. Atoms are subtyped into \textit{objects} $A_o$, \textit{relationships} $A_r$, and \textit{attributes} $A_a$. A \textit{compound} $C$ is defined as a primitive composition of multiple atoms, which corresponds to connections between scene graph nodes. Visual concepts admit two compound types: the attachment of attribute to objects (``black dog”) $C_{ao}$, and the attachment of two objects via a relationship (``man hugs child”) $C_{oro}$.

The composition of these compounds form subgraphs $S$, which can be translated to natural language captions $T$. Conversely, captions $T$ derived from image-text datasets $\mathcal{D}$  can be parsed to become scene graphs $S$.
This extensible language is capable of capturing a number of linguistic phenomena identified in existing literature~\cite{suhr-etal-2019-corpus,parcalabescu-etal-2022-valse}, including the existence of concepts (``a photo with \textit{flowers}''), spatial relationships (``a grill \textit{on the left of} a staircase''), action relationships (``a person \textit{throwing} a frisbee''),  
prepositional attachment (``A bird with \textit{green} wings''), and negation (``There are \textit{no} trucks on the road''). Furthermore, while this study focuses on visual concepts, scene graphs featuring common-sense relationships or other more abstract concepts can be designed; therefore, our methodology is widely applicable~\cite{sap2018atomic}.


\input{tables/dataset_summary}

\subsection{Systematicity}
\label{sec:systematicity}

With our compositional language in place, we now define two dimensions of compositionality—systematicity and productivity—which we adapt to vision-language representations.
\textit{Systematicity} evaluates a model’s ability to systematically recombine seen atoms in compounds. Concretely, 
let $\textsc{Seen}(A, D)$ denote if an atom is seen in a training dataset $\D$,  namely $\exists (I, S) \in \D: A \in S$, and $\textsc{Seen}(C, D)$ denote if a compound is seen in a dataset $\D$,  namely $\exists (I, S) \in \D: C \subseteq S$. To evaluate systematicity, we define three compositional splits: Seen Compounds ($\textsc{SC}$), Unseen Compounds ($\textsc{UC}$) and Unseen Atoms ($\textsc{UA}$).
$\textsc{SC}$ is the split where all compounds (and thus all atoms) of every caption have been seen in the training dataset, \ie $\D_{\textsc{SC}}  = \{(I, S) \in \D_{test} \,|\, \forall C \subseteq S: \textsc{Seen}(C, \D_{train}) \}$. $\textsc{UC}$ is the split where, for each caption, all atoms have been seen but at least one compound has NOT,  \ie $\D_{UC} = \{(I, S) \in \D_{test} \,|\, (\forall A \in S: \textsc{Seen}(A, \D_{train}) \land (\exists C \subseteq S: \neg \mathrm{Seen}(C, \D_{train})) \}$. $\textsc{UA}$ is the split where each caption contains at least one atom that has NOT been seen, \ie $\D_{UA} = \{(I, S) \in \D_{test} \,|\, \exists A \in S: \neg \textsc{Seen}(A, \D_{train})\ \}$. 

\subsection{Productivity} 
\label{sec:prodictivity}
Productivity refers to a capacity to comprehend an unbounded set of expressions. Since the set of atoms in any dataset is finite, a reasonable substitute for testing unbounded comprehension is testing comprehension over increasingly complex scenes. 
Now, an image $I$ does not have a notion of complexity, since it is theoretically infinitely describable; on the other hand, we can define a notion of complexity for a caption $T$: the number of atoms in its corresponding scene graph $|S_T|$. 
\footnote{By avoiding captions with redundant objects (``... a lamb and a lamb and...'') and abstract modifiers (``there are many lampposts''), we ensure atom count is tightly coupled with caption complexity.}
Therefore, a \textit{productive} vision-language model should be able to match a given image to the correct corresponding caption, regardless of that caption's complexity. To evaluate productivity, we define a range of productivity complexity (in our case, $n = 4, 5, \dots, 12$). We need splits of the evaluation dataset based on these complexities, where image-text pairs in a given split have a fixed complexity $n$, and evaluate a model's performance over each split.

\subsection{Compositional evaluation via retrieval}
\label{sec:retrieval}
We evaluate compositional reasoning using zero-shot image-to-text and text-to-image retrieval. This formulation probes the representation space as directly as possible and is already the most common evaluation method for vision-language foundation models~\cite{radford2021learning}. Theoretically, any existing image-text dataset can be used as retrieval sets for our evaluation. However, one challenging limitation in existing datasets renders the metrics evaluated on them inaccurate. Consider using an image query of a ``plant inside a yellow vase on top of a black television.'' Retrieving unintended alternative positives (\eg ``a black television'') is not necessarily incorrect. Similarly, if no other texts in the retrieval set contain a ``plant'' and a ``television'', retrieving the correct text doesn't suggest that the model comprehends the image. 
Ideally, to properly evaluate a model, the retrieval dataset should contain \textit{hard negatives} for every query. A hard negative is a caption that does not faithfully represent the corresponding image, and differs from the ground truth caption by some minimal atomic shift. A example hard negative for the query above is ``\textit{man} inside a yellow vase on top of a black television.'' By erring in a single, granular syntactic or semantic fashion, hard negatives allow for variations in retrieval performance to be attributable to a specific failure mode of a model's compositional comprehension (see Appendix). 
We address this need for a new benchmark dataset to evaluate the systematicity and productivity of vision-language models.



%% file: figures/productivity_figure.tex
\begin{figure*}[t]
     \centering
     \includegraphics[width=\linewidth]{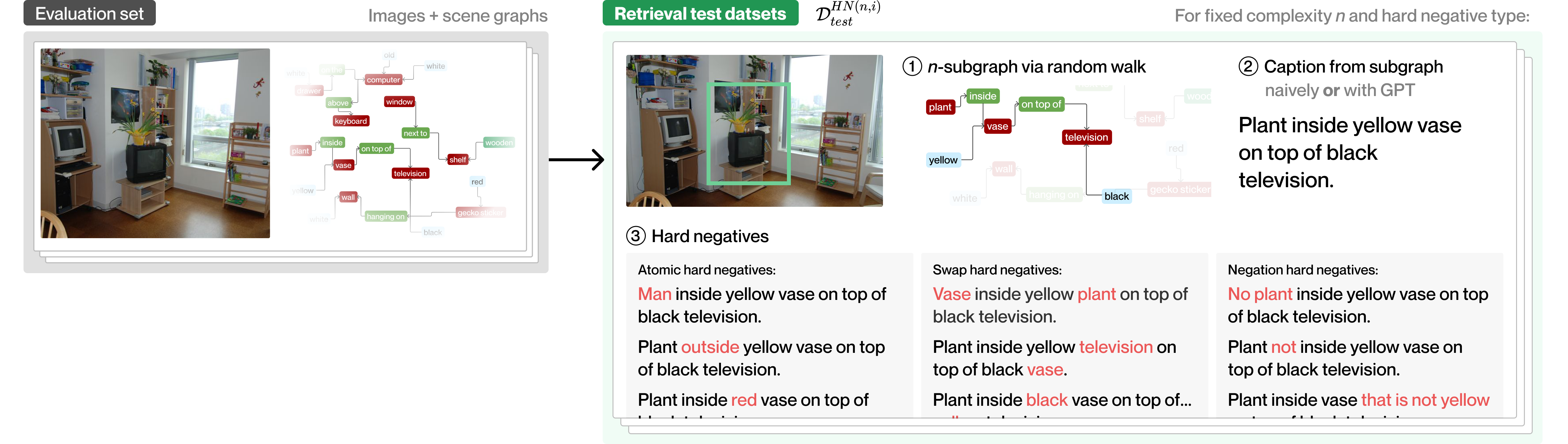}
     \caption{An overview of the \textbf{productivity} retrieval set generation process. By performing random walks on the scene graphs of an evaluation dataset, we generate subgraphs of various complexities. Then, for complexities $n \in \{4, ,5 \dots, 12\}$ and three hard negative types, we populate the retrieval set $\D_{test}^{HN}$ by generating a ground truth caption for each $n$-subgraph and hard negatives for each caption.}
     \label{fig:productivity}
 \end{figure*}

%% file: tables/dataset_summary.tex
\begin{table*}[ht]
\small
\renewcommand{\arraystretch}{1.1}
\centering
\caption{ We summarize the sizes of the eight evaluation datasets we create for systematicity and productivity evaluation.}
\resizebox{\textwidth}{!}{
\begin{tabular}{l rrr p{0.001\textwidth} rrr p{0.001\textwidth} r p{0.001\textwidth} r}
             & \multicolumn{7}{c}{Systematicity}  &                                                     & \multicolumn{3}{c}{Productivity}              \\
             \cmidrule{2-8}\cmidrule{10-12}
             & \multicolumn{3}{c}{$\D_{test}^{RAW}$ (\# of image-text pairs)} & & \multicolumn{3}{c}{$\D_{test}^{HN}$ (\# of texts)} & & $\D_{test}^{RAW}$  && $\D_{test}^{HN}$ \\
             \cmidrule{2-4}\cmidrule{6-8}\cmidrule{10-10}\cmidrule{12-12}
Training data & CC-12M       & YFCC-15M       & LAION-400M       && CC-12M    & YFCC-15M   & LAION-400M   & & Any                        && Any              \\
Dataset size         & 385,777       & 385,777         & 373,703           && 325,523    & 316,668    & 309,342      & & 17,553                      && 183,855          
\end{tabular}
 }
\label{tab:dataset_summary}
\end{table*}

%% file: sections/04_methods.tex
\section{\nameemoji: a large-scale benchmark for vision-language compositionality}

There are several challenges to creating image-text retrieval datasets that evaluate compositional systematicity and productivity. For systematicity, the primary challenge lies in parsing the training dataset for seen atoms and compounds in order to split the data into the three compositional splits. For productivity, the major challenge is generating image-text pairs across different text complexities for the retrieval sets. For both datasets, it is crucial to enumerate different types of hard negatives, and to design an automated hard negative generator which ensures the incorrectness of the negatives it generates. 
We detail our methods for tackling these challenges for future efforts that attempt to create similar benchmarks for other training datasets.

\subsection{Creating systematicity datasets}
To create the three systematicity splits—$\textsc{SC}$, $\textsc{SA}$, $\textsc{UA}$—we parse a given training dataset $\D$ into its constituent atoms and compounds, filter low-quality data, and generate hard negatives (Figure~\ref{fig:systematicity}).


\noindent\textbf{Parsing a dataset into atoms and compounds}
Since we utilize the scene graph representation as our compositional language, we use the Stanford Scene Graph Parser~\cite{schuster2015generating, wu2019unified} to parse texts in $\D_{train}$ into their corresponding scene graphs with objects, attributes and relationships. Since the parser only parses for objects and relationships, we further extract the attributes from the text via spaCy's natural language processing parser by identifying adjective part-of-speech tags. These connected objects, attributes, and relationships constitute our seen atoms and compounds. Similarly, we parse a given $\D_{test}$ and  divide all the image-text pairs into the three splits based on the presence of unseen atoms and/or compounds in the parsed training set. Details on the quality of the scene graph parser can be found in the Appendix.

\noindent\textbf{Filtering low-quality data}
We perform the following filtering steps on the image-text pairs in all splits: we only keep region crops which have an area greater than or equal to $40K$ pixels, occupy at least $10\%$ of the whole image, and whose width-to-height ratio is between $0.5$-$2.0$. We only include text which have at least $2$ atoms and $1$ compound and de-duplicate text using their corresponding scene graphs. 

\noindent\textbf{Generating hard negatives} 
We introduce two types of hard negatives: $\textsc{HN-Atom}$ and $\textsc{HN-Comp}$.
$\textsc{HN-Atom}$ replaces $A_a$, $A_o$, or $A_r$ in the text with an atomic foil. For example, for the caption ``a grill on top of the porch'', one $\textsc{HN-Atom}$ can be ``a grill \textit{underneath} the porch'', where the $A_r$ ``on top of'' is replaced by ``underneath''. Since captions and scene graphs are not exhaustive, this replacement must be done carefully. For example, if a dog is white and furry, but only ``white'' is annotated, replacing the atom ``white'' with ``furry'' would result in a correct caption. To minimize errors, we employ WordNet~\cite{miller1995wordnet} to pick replacement atoms that are either antonyms (``\textit{black} dog'') or share the same grand-hypernym (``\textit{pink} dog") with respect to the original atom. 
Furthermore, we use BERT to select the most sensical negatives for each ground truth caption \cite{devlin2018bert, parcalabescu-etal-2022-valse}. 
$\textsc{HN-Comp}$ concatenates two compound foils  where each contains an atomic foil. For instance, one $\textsc{HN-Comp}$ of the caption ``a pink car'' can be ``a \textit{blue} car and a pink \textit{toy}'', where ``blue'' and ``toy'' are the atomic foils in the two compounds foils ``blue car'' and ``pink toy''. We only generate negatives for one-compound examples for systematicity evaluation, as productivity covers complex captions with more atoms. 


\input{figures/syst_hn_recall_at_1}

\subsection{Creating productivity datasets}
We first generate ground truth captions for scene graphs of varying complexity, filter for data quality, and then generate hard negatives for each example (Figure~\ref{fig:productivity}).

\noindent\textbf{Generating captions}
We systematically generate captions of different atom counts for each image.
Given a scene graph, we perform a random walk of length $n$ through the graph to generate a subgraph.
Each subgraph corresponds to a specific region of the image, determined by the union of the bounding boxes of the subgraph atoms. We filter out low-quality regions using the same process as systematicity with additional deduplication on patches that overlap by $\geq 75\%$. 
For simple subgraphs ($n=4$), we produce captions using handcrafted templates. 
For larger subgraphs ($n \geq 5$), we leverage GPT-3~\cite{brown2020language} (text-davinci-002) to generate captions based on a text description of the scene graph, which lists all objects and relationships. 
We prompt GPT-3 using 5 manually written captions per complexity, filtering out captions where GPT-3 errs and omits atoms from the subgraph during generation (see 
 more details in Appendix).


\noindent\textbf{Generating hard negatives}
For productivity, we employ three hard negatives types ($\textsc{HN-Atom}$ from systematicity, $\textsc{HN-Swap}$, and $\textsc{HN-Neg}$) corresponding to three hypothesized model error modes.
First, as a caption's complexity increases, a model may begin to ignore individual atoms. $\textsc{HN-Atom}$ randomly selects an atom from the caption and replaces it with an incorrect atom.
Second, as a caption's complexity increases, a model may treat captions as ``bags of words'', ignoring syntactic connections built out of word order. A \textit{swap hard negative} ($\textsc{HN-Swap}$) accordingly permutes atoms of the same subtype in a caption. This hard negative is similar to Winoground~\cite{thrush2022winoground}, but in the context of varying caption complexity. On top of Wordnet, we use entailment with RoBERTa to further filter errant \textsc{HN-Swap} hard negatives \cite{liu2019roberta}.
Finally, as a caption's complexity increases, a model may begin to lose comprehension of negations. A \textit{negation hard negative} ($\textsc{HN-Neg}$) either negates the entire caption or a specific atom. Refer to the Appendix for details on generating $\textsc{HN-Swap}$ and $\textsc{HN-Neg}$.


\input{figures/produtivity_hard_negs_plot}
\input{figures/productivity_hard_negs_plot_other_models}
\subsection{The final benchmark datasets}
For both productivity and systematicity, we generate two test datasets:
$\D_{test}^{HN}$, which contains image-ground truth text pairs along with all generated hard negatives, and $D_{test}^{RAW}$, which contains only image-ground truth text pairs. 
To measure the data quality, we randomly sample $2\%$ of productivity ground truth captions generated by GPT-3 and $1\%$ of the queries in the productivity and systematicity $\D_{test}^{HN}$ sets for manual human verification. We assign $2$ annotators to each set and measure both generated quality and intra-annotator agreement.
$87.9\%$ of sampled productivity ground truth captions generated by GPT-3 are rated as faithful to the image, with an average pairwise annotator agreement of $88.8\%$. 
$83.7\%$ of productivity and $86.0\%$ of systematicity hard negatives were rated as genuine negatives (\ie made factually incorrect statements about the image), with pairwise annotator agreements of $84.3\%$ and $83.7\%$ respectively. 

%% file: figures/syst_hn_recall_at_1.tex
\begin{figure*}[t]
    \centering
    \includegraphics[width=\linewidth]{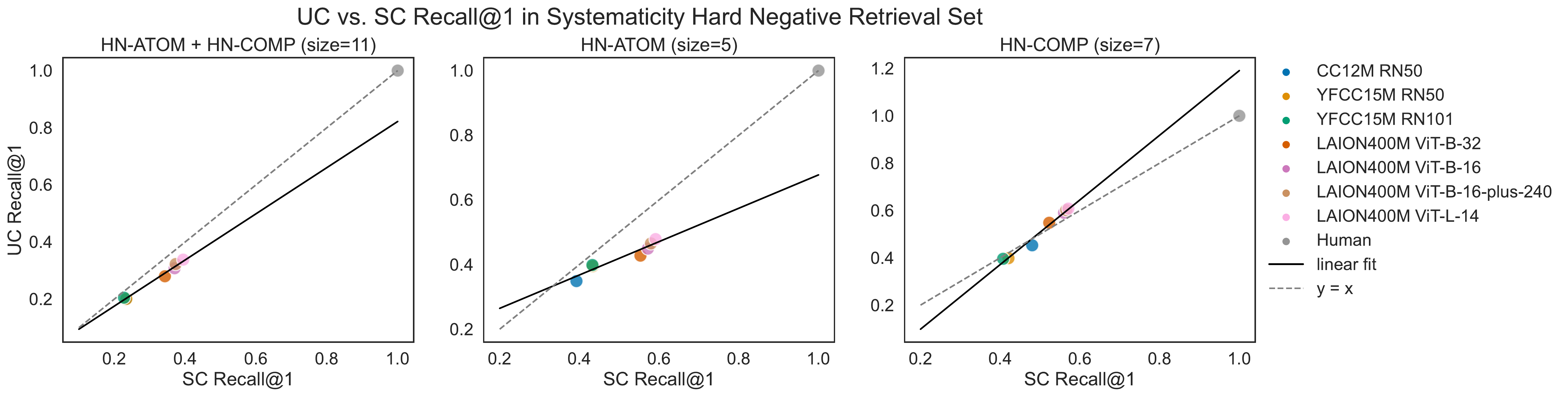}
    \caption{\textit{Systematicity analysis.} We plot models' recall@1 on the Seen Compounds vs. Unseen Compounds split of the systematicity retrieval set with hard negatives \textsc{HN-Atom}, \textsc{HN-Comp} and both types. We observe a consistent drop in models' performance from the SC to UC split when the retrieval set contains \textsc{HN-Atom} or both types, and little to no difference when it contains only \textsc{HN-Comp}.}
    \label{fig:syst_hn_recall_at_1}
\end{figure*}

%% file: figures/produtivity_hard_negs_plot.tex
\begin{figure*}[hbt]
     \centering
     \includegraphics[width=0.8\linewidth]{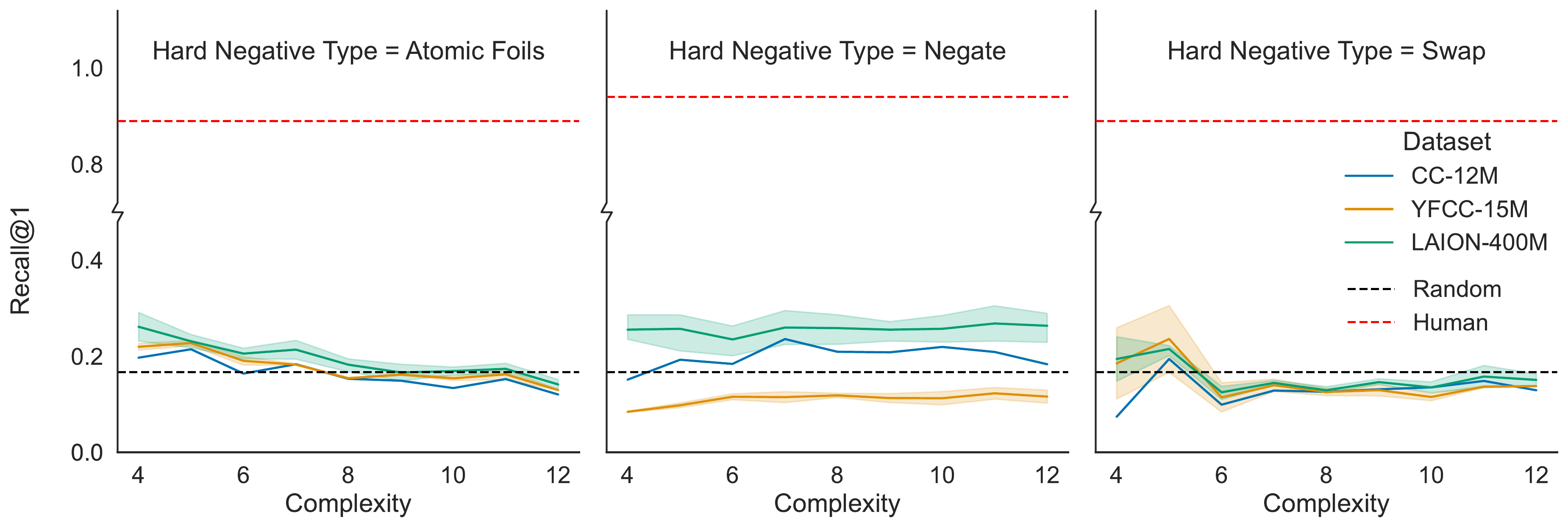}
     \caption{\textit{Productivity Analysis.} We plot models' Recall@1 on the hard negatives retrieval set against complexity, averaged across all models pretrained on all three training datasets. We find that models' ability to retrieve the ground-truth degrades as complexity increases.}
     \label{fig:prod_hard_negs}
 \end{figure*}
 

%% file: figures/productivity_hard_negs_plot_other_models.tex
\begin{figure*}[hbt]
     \centering
     \includegraphics[width=0.8\linewidth]{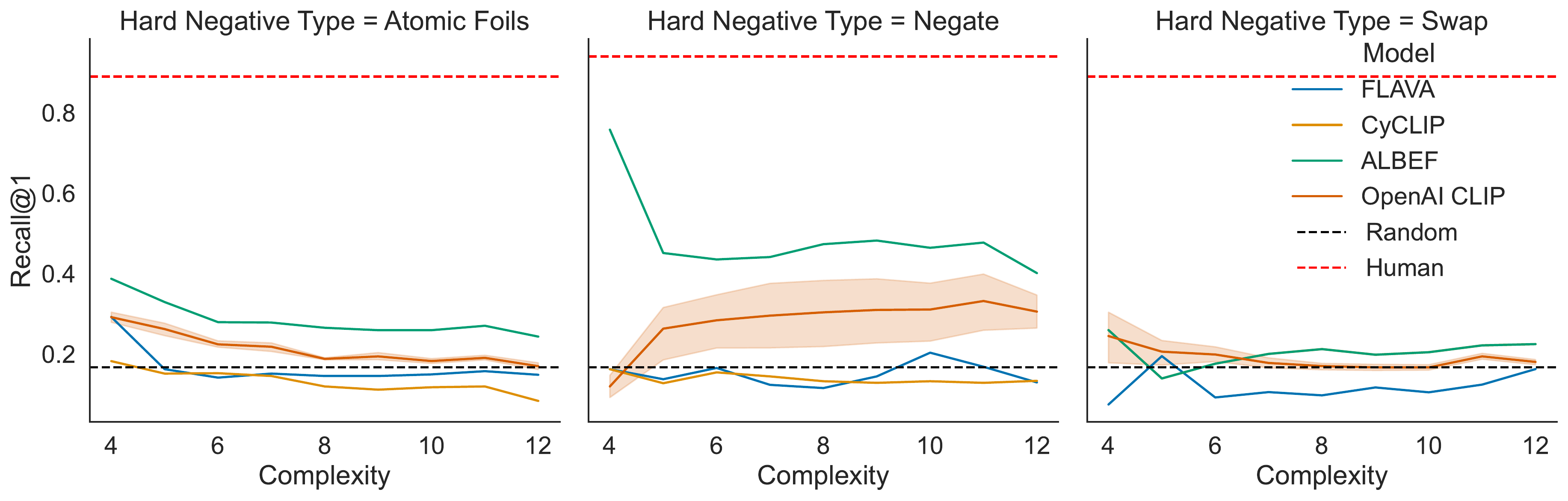}
     \caption{\textit{Productivity Analysis on Additional Foundation Vision-language Models.} We plot models' Recall@1 on the productivity hard negatives retrieval set against complexity, where OpenAI CLIP's performance is averaged across five models RN50, RN101, ViT-B-16, ViT-B-32 and ViT-L-14. We find that all models' retrieval performance decreases as complexity increases in both the \textsc{HN-ATOM} and \textsc{HN-SWAP} retrieval sets. For the \textsc{HN-NEG} set, all models except for CLIP either drop in performance or remain at random chance.}
     \label{fig:prod_hard_negs_other_models}
 \end{figure*}
 

%% file: sections/05_experiments.tex
\section{Experiments}
We present our experimental setup and results with six takeaways. First, our systematicity experiments show performance decreases consistently on compounds unseen in training. Second, the greatest drop between splits occurs for models trained on LAION-400M. Third, our productivity results reveal models' retrieval performance decays with increasing complexity. Fourth, we find that dataset size has no impact on compositionality. Fifth, we find no clear trend relating model size to compositionality. Finally, models' zero-shot ImageNet classification accuracy correlates with retrieval performance on the systematicity dataset but not systematic generalization to the UC split or productivity.

\noindent\textbf{Datasets.} 
We utilize Visual Genome to create our test datasets. For systematicity, image patches and corresponding spelling-corrected region descriptions are used.
We provide three different splits for $\D_{test}^{HN}$, for three training datasets: CC-12M, YFCC-15M and LAION-400M. For productivity, Visual Genome's image-scene graph pairs are used to create captions and hard negatives for $\D_{test}^{RAW}$ and $\D_{test}^{HN}$ (Table \ref{tab:dataset_summary}). 


\noindent\textbf{Models.} We firstly evaluate seven vision-language models pretrained with contrastive loss~\cite{sohn2016improved} across three commonly used image-text datasets: Conceptual Captions 12M (CC-12M)~\cite{changpinyo2021conceptual}, a subset of the YFCC100M dataset (YFCC-15M)~\cite{thomee2016yfcc100m, radford2021learning} and LAION-400M~\cite{schuhmann2021laion}. We limit our evaluation to models openly released in the OpenCLIP repository~\cite{Ilharco_OpenCLIP_2021} for systematicity evaluation. 
These include ResNet (RN)~\cite{he2016deep} and Vision Transformer (ViT)~\cite{dosovitskiy2021an} encoders of different sizes: RN50, RN101, ViT-B-16, ViT-B-16-plus-240, ViT-B-32 and ViT-L-14. Additionally, since productivity evaluation is not restricted to models that were trained on publicly released datasets, we conduct productivity evaluation on other foundation vision-language models as well. Specifically, we consider OpenAI's CLIP~\cite{radford2021learning} with ResNet and ViT backbones, CyCLIP~\cite{https://doi.org/10.48550/arxiv.2205.14459} (a variant of CLIP introducing auxiliary losses that regularize the gap in similarity scores between mismatched pairs, trained on Conceptual Captions 3M~\cite{sharma2018conceptual} with a ResNet-50~\cite{he2016deep} backbone), ALBEF~\cite{li2021align} (additionally trained with a masked language modeling and image-text matching loss) and FLAVA~\cite{https://doi.org/10.48550/arxiv.2112.04482} (which further adds unimodal losses for image and text domains).

\noindent\textbf{Retrieval.} For $\D_{test}^{HN}$, we perform image-to-text retrieval and stratify results by split and hard negative type. For systematicity, the splits are $SC$, $UC$, and $UA$; for productivity, the splits are by caption complexity $n$ (denoted $\D_{test}^{HN, n}$). Each retrieval task is between one image and its ground truth caption plus $h$ hard negatives of a single type (see Appendix).
We adopt commonly used retrieval metrics Recall@1, 3, 5 and Average Recall@K. For $\D_{test}^{RAW}$, retrieval experiments are described in the Appendix.

\input{figures/imagenet_correlation_syst.tex}
\input{figures/imagenet_correlation_prod.tex}
\subsection{Systematicity evaluation}
\paragraph{Model performance on the $\D_{test}^{HN}$ dataset for systematicity decreases monotonically when compounds are unseen.} 

We first observe a monotonic decrease in recall@1 from the Seen Compounds to the Unseen Compounds split on the systematicity $\D_{test}^{HN}$ set consisting of both \textsc{HN-Atom} and \textsc{HN-Comp} (Figure \ref{fig:syst_hn_recall_at_1} left). This drop is relatively small ($2-4\%$) for the CC-12M and YFCC-15M trained models and the most pronounced for models trained on the largest dataset LAION-400M~\cite{schuhmann2021laion}, with the decrease reaching $6\%$ for the ViT-B-32 model. However, CC-12M and YFCC-15M models also significantly underperform LAION-400M models in general, meaning that small drops between sets may be due to overall poor performance rather than improved systematic generalization. In comparison, human oracle experiments generalize with $100\%$ accuracy to $\D^{HN}_{test}$.

Similar to the overall results, there is also a consistent discrepancy between the SC and UC split on the $\D_{test}^{HN}$ subset consisting of \textsc{HN-Atom} only (Figure \ref{fig:syst_hn_recall_at_1} center). This drop is consistently smaller ($3-5\%$) for models trained on CC-12M and YFCC-15M, but pronounced (higher than $10\%$, reaching $12.5\%$ drop for ViT-B-32) for LAION-400M models.

On the \textsc{HN-Comp} subset (Figure \ref{fig:syst_hn_recall_at_1} right), we find little ($1-3\%$) to no difference in performance between the two splits. We hypothesize that this is due to the lower difficulty of the \textsc{HN-Comp} hard negatives, as they introduce more foils to the caption, are always longer than the ground truth, and thus offer more opportunities for the model to correctly distinguish the ground truth. This hypothesis is supported by the fact that Recall@1 values on \textsc{HN-Comp} are overall higher than the ones on \textsc{HN-Atom} even though the \textsc{HN-Comp} retrieval set size is larger than that of \textsc{HN-Atom}.



\subsection{Productivity evaluation}
\paragraph{Models' performance decreases with complexity on $\textsc{HN-Atom}$ and $\textsc{HN-Swap}$ negatives.} 
At small complexities such as $n=4$, we observe that model retrieval  quality is well above random chance  (Figure \ref{fig:prod_hard_negs}). However, as caption complexity increases, we observe a steady decrease in performance, nearing random chance for $\textsc{HN-Atom}$ and dipping below it for $\textsc{HN-Swap}$ negatives. 
Similarly, we find that the same downward trend persists for other vision-language foundation models (Figure~\ref{fig:prod_hard_negs_other_models}) on $\textsc{HN-Atom}$, and these models also perform near random chance on $\textsc{HN-Swap}$. Importantly, the downward trend occurs for FLAVA and ALBEF even though their training set contains Visual Genome images. We note that for \textsc{HN-Neg} negatives, the OpenAI CLIP models do not adhere to the downward trend, achieving their lowest scores for the lowest complexity. Their performances on higher complexities, however, show great variation. Overall, we find that all vision-language foundation models in our evaluation struggle at the productivity hard negative retrieval sets, demonstrating near-random chance performance and/or worse performance at higher caption complexity. 

\noindent\textbf{We see no effect of dataset size on productivity.} We do not observe a clear advantage for larger pretraining datasets in our productivity evaluation.
For atomic and swapping foils, we see similar performance for models trained on the three datasets, with slightly worse performance on atomic foils for the CC-12M trained models. 
However, on negation hard negatives (Figure \ref{fig:prod_hard_negs}), we see variable performance across training sets, with CC-12M models outperforming larger models trained on larger datasets YFCC and LAION.



\subsection{Effect of model size}
\noindent\textbf{We find no trends relating compositionality to model size.}
Overall, we note that the LAION trained models (which are both larger models and trained on larger datasets) achieve \textit{significantly} better absolute performances than smaller models. However, model's systematicity and productivity remain indifferent to the size of the model itself (Figures~\ref{fig:syst_hn_recall_at_1} and~\ref{fig:prod_hard_negs}).





\subsection{Correlation with ImageNet performance}
\noindent\textbf{We find that zero-shot ImageNet accuracy strongly correlates with models' Recall@1 on the hard negative retrieval sets except for productivity $\textsc{HN-SWAP}$.}
Specifically, we acquire $R^2$ scores of $0.95, 0.80$ for the systematicity SC and UC split on \textsc{HN-ATOM}, and $0.91, 0.97$ on \textsc{HN-COMP} (Figure~\ref{fig:correlation_systematicity}). On productivity datasets, we obtain $R^2$ scores of $0.60, 0.79, 0.55$ for $\textsc{HN-Atom}$, $0.92, 0.78, 0.88$ for $\textsc{HN-Neg}$ and $0.07, 0.08, 0.47$ for $\textsc{HN-Swap}$ negatives on complexity $n=4,8,12$ respectively (Figure~\ref{fig:correlation_productivity}). 
However, this correlation does not imply that models' zero-shot ImageNet performance correlates with systematic or productive generalization, which is indicated by small or no \textit{difference} between the $SC$ and $UC$ and complexity splits.  





%% file: figures/imagenet_correlation_syst.tex
\begin{figure}[t]
    \centering
    \includegraphics[width=\columnwidth]{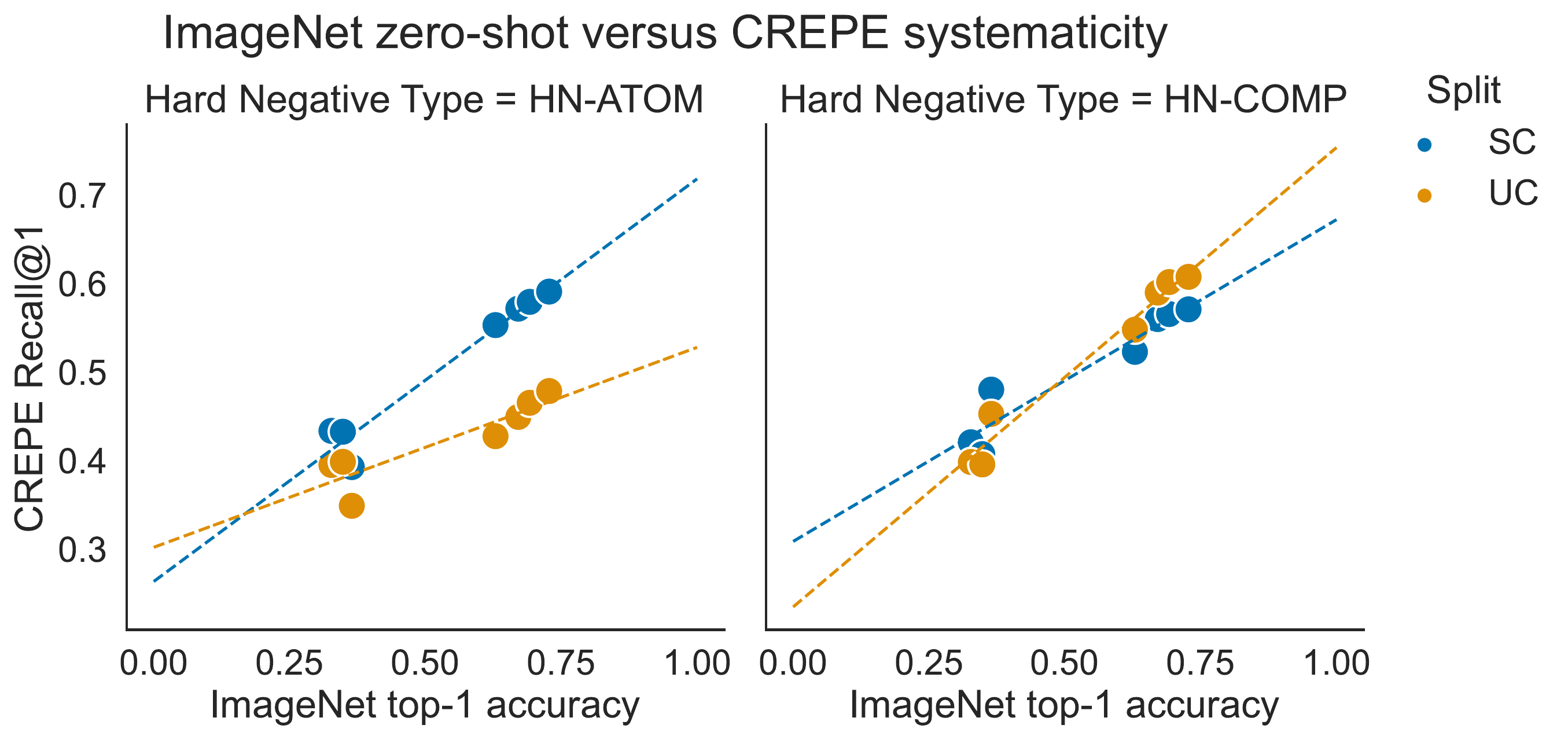}
    \caption{A plot of ImageNet 
 zero-shot top-1 accuracy vs. Recall@1 on CREPE's systematicity hard-negative sets. We observe a strong correlation for both splits on \textsc{HN-ATOM} and \textsc{HN-COMP}.}
    \label{fig:correlation_systematicity}
\end{figure}

%% file: figures/imagenet_correlation_prod.tex
\begin{figure*}[ht!]
    \centering
    \includegraphics[width=0.92\textwidth]{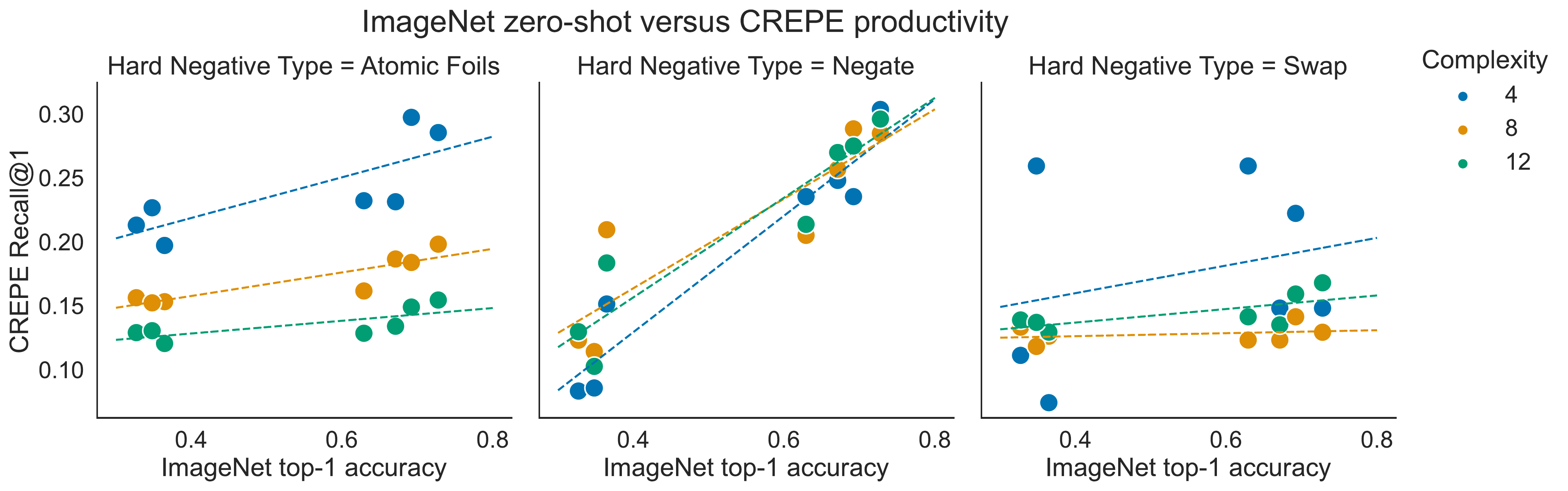}
    \caption{A plot showing the correlation between zero-shot top-1 accuracy on ImageNet and Recall@1 on CREPE's productivity hard negative sets for complexities of $4, 8$, and $12$. Overall, we find a strong correlation between ImageNet accuracy and Recall@1 on our productivity $\textsc{HN-Neg}$ ($R^2 > 0.78$) and $\textsc{HN-Atom}$ ($R^2 > 0.54$) sets and weak to no correlation on $\textsc{HN-Swap}$. } 
    \label{fig:correlation_productivity}
\end{figure*}

%% file: sections/06_discussion.tex
\section{Discussion}

\paragraph{Limitations.} 
First, although our data validation protocols verified our generated hard negatives for productivity as high-quality, approximately $70\%$ of \textsc{HN-Swap} and of \textsc{HN-Neg} negatives were rated as correct. While this does not invalidate our key productivity result, this noise is a limitation of \name{} and could hinder future evaluations once foundation models begin performing better. Second, our evaluation only covers a limited set of vision-language foundation models that were trained with contrastive loss. Additionally, given the computational requirements associated with training a foundation model, our experiments centered around model architectures that were already available publicly. We hope that future foundation models are evaluated with our publicly available \name{} benchmark. Third, while we observe text-to-image and image-to-text retrieval to have similar trends for our systematicity experiments, we lack text-to-image datasets with hard negatives. Future work can explore mechanisms to generate counterfactual negative images.

\paragraph{Conclusion.}
We present \nameemoji{}, a collection of image-to-text retrieval datasets with hard negative texts for evaluating pretrained vision-language models' systematicity and productivity. We demonstrate that models struggle with compositionality along both axes, with performance drops across compositional splits and complexities.
We expect that \name{} will provide a more systematic evaluation 
to benchmark the emergence of compositionality as future models improve.
Finally, 
researchers can leverage our hard-negative generation method to create training batches with hard negatives to improve vision-language compositionality.

%% file: sections/07_conclusion.tex

%% file: sections/08_appendix.tex



\section{Additional details on dataset generation}

\input{tables/productivity_error_modes.tex}

\subsection{Hard negative types}
In both our productivity and systematicity experiments, we rely on hard negatives to ensure that the retrieval sets we construct meaningfully probe a model's comprehension. Specifically, to \textit{granularly} probe a model's comprehension, we identify a set of common failure modes of non-compositional models and design hard negative types that address each of these failure modes. Examples of each failure mode and hard negative type are outlined in Table \ref{tab:hard_negatives}. 

\subsection{Scene graph parser verification}
\input{tables/scene_graph_parser.tex}

To generate data splits for our systematicity experiments, we employed a rule-based implementation of the Stanford Scene Graph Parser ~\cite{schuster2015generating, wu2019unified}. To verify its performance, we randomly sample 20 captions from each of CC-12M, YFCC-15M and LAION-400M and manually annotate scene graphs for the captions. We report precision and recall values for object, attribute and relationship atoms and object-relationship-object triplets on Table \ref{tab:scene_graph_parser}. For CC-12M, YFCC-15M and LAION-400M, the object precision was $88.14, 96.24, 70.00\%$, attribute precision was $93.00, 94.44, 72.22\%$ and triplet precision was $91.67, 92.31$ and $87.00\%$ respectively. For recall values, on the other hand, we found that object recall was $83.06, 93.33, 60.68\%$, attribute recall was $56.51, 75.56, 36.11\%$ and triplet recall was $64.04, 81.11$ and $39.55\%$ respectively. The precision values help determine whether the atoms the parser identifies are valid, while the recall values help determine whether the parser \textit{can} identify the atoms and triplets present in the caption, important for the validity of our seen compounds (SC) and unseen compounds (UC) splits.

We find that the parser's precision values are high throughout for each dataset. Recall values are lower compared to precision, particularly for the LAION dataset, where captions can be more similar to bags of words rather than well structured sentences. We note, however, that if compounds were incorrectly placed into the UC set due to poor recall, our systematicity task would become easier. As all models experience drops in performance between SC and UC splits, we do not observe this.



\subsection{Productivity caption generation}
As discussed in the main text, each instance in the productivity test dataset is a image-text pair of complexity $n$ with a set of hard negative captions.
To generate such examples, we begin by sampling a $n$-node subgraph from a scene graph in Visual Genome~\cite{krishnavisualgenome}. We sample this subgraph using a random walk (see the paragraph titled \textbf{Random walk}). This subgraph is then transformed into a caption either using a template or GPT-3 (see the paragraph titled \textbf{Caption generation}).
Finally, we crop the original image to the union of all object bounding boxes in the subgraph (see main text). We describe these details below.

\paragraph{Random walk} Given a scene graph $G$, we generate an $n$-atom subgraph ($n \leq |G|$). We initialize a subgraph $S$ with a single random object in $G$. While this subgraph contains less than $n$ atoms, a compound $C$ consisting of at least one unadded atom is added to $S$. If $C$ is a relationship compound ($C_{oro}$), the walk continues from the newly added object; otherwise, the walk is continued from the same object. If the entire connected component of the scene graph is exhausted, another object is selected at random from a different connected component. This process ends when $n$ atoms are added to the subgraph. We discard all walks that result in insufficient number of atom.

\begin{figure*}[th]
    \centering
    \includegraphics[width=\textwidth]{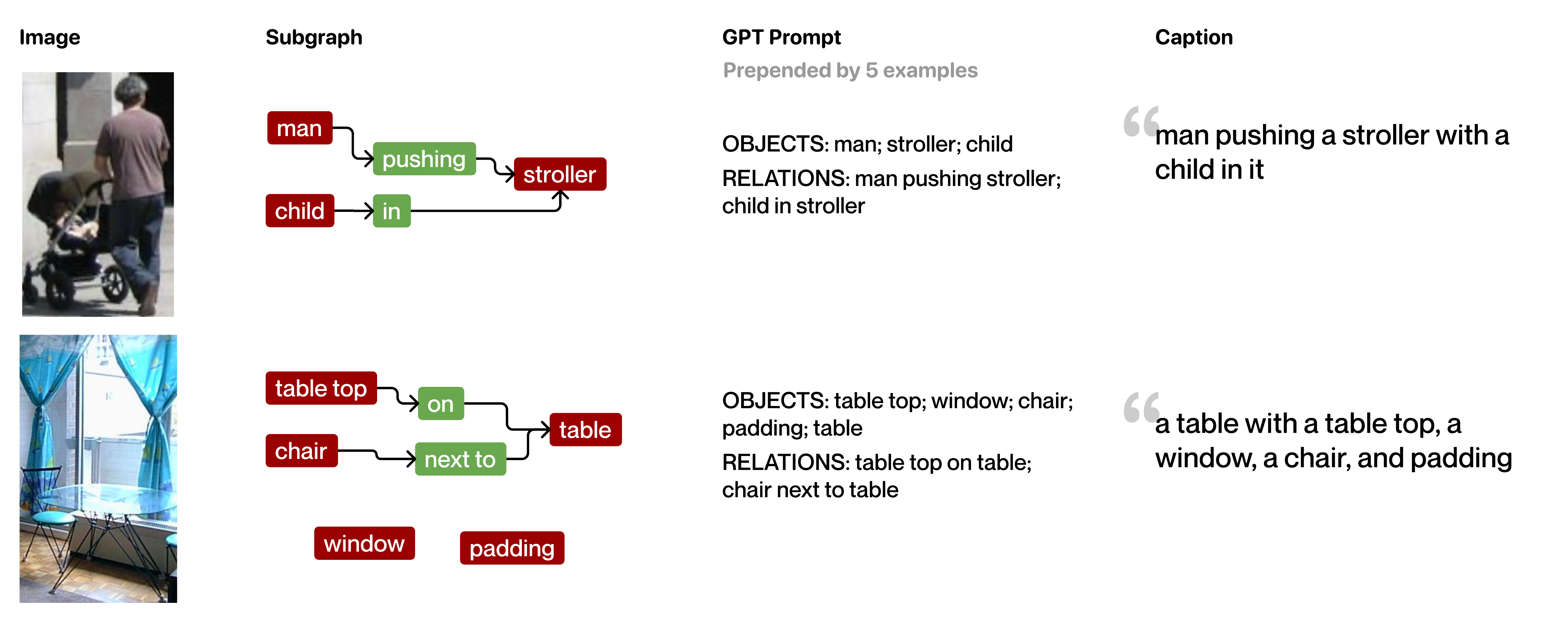}
    \caption{Examples of image-text pairs we generate for our productivity evaluation. The image is the union of the objects bounding boxes in the subgraphs. We also showcase the GPT-3 prompt associated with the subgraph and their corresponding generated ground-truth captions.}
    \label{fig:prod_caption_examples}
\end{figure*}

\paragraph{Caption generation} To generate captions, we either utilize hand crafted templates or use GPT-3. For subgraphs of complexity $n > 4$, we use GPT-3 to generate a coherent caption for each prompt; otherwise, we use the templates.
When prompting GPT-3 to produce captions, we populate the the first line of the prompt with a list the objects in the subgraph, prepended with their attributes. If multiple instances of an object type occur (\eg, we have two objects both with name ``window'' in the graph), we append a numerical suffix to distinguish between then (\eg ``window1'' from ``window2.'').
On the second line of the prompt, we list all the relationships between objects in the graph, in the form \texttt{subject relationship object}.
Additionally, we manually generate 5 caption examples per complexity from random subgraphs and prepend both the random subgraph and the manually generated caption to the prompt above, as few-shot training examples for GPT-3.
We provide examples of graphs, prompts, and their generated captions in Figure \ref{fig:prod_caption_examples}.

For examples of complexity $n=4$, we found that stringing together a simple templated prompt was sufficient to produce fluent captions. This was done by prepending attributes in front of objects and stringing together subjects, relations, and objects in the correct order.
For example, a subgraph containing \texttt{boy=(tall,blue); grass=(green); (boy, on, grass)} would be templated as \texttt{tall and blue boy on green grass}. Any disconnected atoms are appended with the prefix ``and a.''

\paragraph{Data verification.} 
\input{tables/productivity_gt_annotation_breakdown}
We manually verify the accuracy of our produced productivity dataset. We provide a breakdown of annotators' scores for GPT-3 caption faithfulness across complex subgraphs with $n \ge 7$ in Table \ref{tab:manual_eval_gt_complexity}. 
We see that scores are consistently high for ground-truth captions across complexities.

\subsection{Hard negative generation details}
We provide additional detail for the procedure of generating a hard negative of types $\textsc{HN-Swap}$ and $\textsc{HN-Neg}$. Suppose throughout that for a given image and its annotated scene graph $G$, we seek to generate a hard negative caption for caption $t$ associated with the subgraph $S \subseteq G$.
\label{sec:swap_atoms}
\paragraph{\textsc{HN-Swap}} The following pairs of atoms could be swapped to create a hard negative for $S$:
    \begin{itemize}
        \item The subject $A_o$ and object $A_o'$ of a relationship compound $C_{oro} \in S$.
        \item Two attributes $A_a$ and $A_a'$ attached to distinct objects $A_o$, $A_o'$, such that one attribute is not present for the other object in $G$ and vice versa. ($(A_a, A_o') \not\in G$ and $(A_a', A_o) \not\in G$).
        \item Two objects $A_o$, $A_o'$ not connected by a relationship such that their swapping within $G$ does not create an identical graph.
    \end{itemize}
    Additionally, some swap hard negatives generated are \textit{permutations} rather than a swapped pair:
    \begin{itemize}
        \item One attribute $A_a$ can be transferred from one object $A_o$ to another object $A_o'$, so long as that attribute doesn't apply to the new object ($(A_a, A_o') \notin G$).
        \item For low complexities ($n = 4$), any permutation of atoms of the same type are allowed. For example: (``There is a dog on the bed and also a nightstand" $\rightarrow$ ``There is a \textit{nightstand} on the \textit{dog} and also a \textit{bed}") 
    \end{itemize}

\paragraph{\textsc{HN-Neg}} We verify with $G$ to ensure that negating an atom results in an incorrect caption. If an attribute $A_a$ connected with $A_o$ is negated,  we ensure that there does not exist an object of $A_o'$ that doesn't have an attribute $A_a$ but shares all the other attributes of $A_o$. For example, if we negate ``black" in ``Black dog on a building", we ensure there doesn't exist another dog on the building that isn't black. Similar checks are performed for negating relationships and objects. When a relationship $A_r$ connecting $A_o$ and $A_o'$ is negated, there cannot exist another identical subject and object pair connected by a different relationship $A_r'$. When an object is negated, there cannot exist any other object with the same attributes and relationships.

\subsection{Test dataset sizes, examples, and additional verification}
\input{tables/image-text-counts}
\input{tables/retrieval_set_size}
\input{figures/syst-dataset-viz.tex}
\input{figures/prod-dataset-viz.tex}
Table \ref{tab:image_text_counts} expands on Table 1 from the main paper to provide a breakdown of the number of image-text pair per hard negative type and, for productivity, for each sentence complexity. We remark that $\D^{RAW}_{test}$, which contains only image-ground-truth caption pairs, is a \textit{superset} of the ground-truth captions in $\D^{HN}_{test}$. This is because, for some ground truth captions in $\D^{RAW}_{test}$, a sufficient number of hard negatives to perform retrieval in $\D^{HN}_{test}$ could not be generated. Additionally, due to the prevalence of rare atoms, we could only generate valid hard negatives for very few captions in the UA split. Therefore, we omit the evaluation on the UA split with hard negatives and focus on the analysis of results between the SC and UC split, which is more interesting as models have seen all the atoms in both splits. Table~\ref{tab:retrieval_set_size} summarizes the text retrieval set size of each image query for both $\D_{test}^{raw}$ and $\D_{test}^{HN}$ in our systematicity and productivity evaluation. Figures \ref{fig:syst_viz} and \ref{fig:prod_viz} present examples of ground truth captions and hard negative captions in our test datasets for systematicity and productivity, respectively.

We provide a breakdown of annotators' scores for the accuracy of productivity hard negatives in Table \ref{tab:manual_eval_prod_hard_negs}.  A hard negative caption is accurate if it contains incorrect facts about the image. We find that the accuracy and pairwise agreement of the \textsc{HN-Atom} is the highest and much higher than those of \textsc{HN-Swap} and \textsc{HN-Neg}.
\input{tables/productivity_hard_negs_annotation_breakdown}

\subsection{Systematicity hard negative dataset details}
\input{tables/atom_comp_counts}
\input{figures/syst_atom_frequency}
Table~\ref{tab:atom_comp_counts} summarizes the number of unique atoms and compounds in the SC and UC split of the systematicity hard negative set. Additionally, we plot the atom count in the systematicity test set vs. the training set (on a log scale). As shown in Figure~\ref{fig:syst_atom_frequency}, we see that the atom count in the training set is always on the same scale across both splits for the same training dataset (x-axis in each row). We further observe that the atom distributions are similar in the SC and UC splits. These suggest that the atoms appearing in the UC split are not substantially rarer or more difficult than the ones in the SC split.

\section{Additional evaluation results}

\label{sec:full-results}
\subsection{Full retrieval results on hard negative datasets}
\input{tables/systematicity_hard_negs}
\input{tables/systematicity_hard_negs_atom}
\input{tables/systematicity_hard_negs_comp}
\paragraph{Systematicity}
We additionally include the full retrieval results on $\D_{test}^{HN}$ with both \textsc{HN-Atom} and \textsc{HN-Comp}, \textsc{HN-Atom} only and \textsc{HN-Comp} only in Tables~\ref{tab:systematicity_hard_negs},~\ref{tab:systematicity_hard_negs_atom} and~\ref{tab:systematicity_hard_negs_comp}. We note that as we relax the metric from R@1 to R@3, the difference between models' performance in the SC and UC split decreases. 

\paragraph{Productivity}
\input{tables/productivity_hard_negs_atom}
\input{tables/productivity_hard_negs_swap}
\input{tables/productivity_hard_negs_negate}
We report the full retrieval results on productivity $\D_{test}^{HN}$ sets in Tables~\ref{tab:productivity_hard_negs_atom},~\ref{tab:productivity_hard_negs_negate} and~\ref{tab:productivity_hard_negs_swap}. 

\label{sec:raw-results}
\subsection{Retrieval results on raw datasets}
\input{tables/systematicity_weak_negs}
\input{tables/productivity_weak_negs}
In addition to $\D^{HN}_{test}$ retrieval experiments, we perform retrieval experiments with $\D^{RAW}_{test}$. 

We perform \textit{both} image-to-text and text-to-image retrieval within splits of $\D_{test}^{RAW}$. Each retrieval task is between one image and every caption in the split, or vice versa. We report the mean and standard deviation of Recall@1 across K-fold retrievals (where $K = \min(20, \lfloor \frac{|\D^{RAW}_{test}|}{N} \rfloor)$, and $N = \min\{|SC|, |UC|, |UA|\} = 1855$ for systematicity and $N = \min_{n\in\{4 \ldots 12\}}|\D_{test}^{RAW, n}| = 1508$ for productivity), as the data size varies across compositional splits and complexities.

\paragraph{Systematicity} 
We present the systematicity retrieval results on $\D_{test}^{raw}$ in Table~\ref{tab:systematicity_weak_negs}, where each retrieval set for an image consists of the captions of the other images. We continue to observe a monotonic decrease in performance when compounds are unseen. Additionally, we continue to observe a drop in performance for larger training datasets. In particular, we see a similar drop in performance for LAION-trained models across both the image-to-text and text-to-image tasks. We also observe larger drops on LAION-trained models than for $\D_{test}^{HN}$ when moving across the $\textsc{SC}\to \textsc{UC}$, and across $\textsc{UC}\to \textsc{UA}$ splits, with LAION models dropping as much as $13\%$ for ViT-L/14.

\paragraph{Productivity} We additionally present the productivity retrieval results on $\D_{test}^{raw}$ in Table~\ref{tab:productivity_weak_negs}. We observe that models' Recall@1 generally increases as the caption complexity increases. We hypothesize that models' low performance in the low-complexity subset is caused by false negatives in the original dataset: since the captions are simple and likely true for multiple images, there are multiple false negatives in the retrieval set, making these numbers unreliable. As the captions become more complex, however, the chance of such false negatives is lower. This means there are more true negatives in the higher-complexity subsets, making retrieval easier for these models.

\input{figures/productivity_hard_negs_overall}
\subsection{Retrieval results with all hard negatives at once}

\paragraph{Productivity}
We present models' retrieval performances over the whole productivity $\D_{test}^{HN}$ dataset, where each retrieval set contains one ground truth caption and fifteen hard negatives, five for each of the three types \textsc{HN-Atom}, \textsc{HN-Swap} and \textsc{HN-Neg}. We find in Figure \ref{fig:prod_hard_negs_overall} that models' Recall@1 performance decreases with complexity, which aligns with the findings on the separate retrieval sets for \textsc{HN-Atom}, \textsc{HN-Swap} and \textsc{HN-Neg}.


\input{tables/systematicity_qualitative_analysis}
\subsection{Qualitative analysis on systematicity evaluation}
We perform a qualitative analysis to better understand why the LAION-400M trained models ViT-B-16 and ViT-L-14 show a large versus small performance drop from the Seen to Unseen Compounds split respectively.  Table \ref{tab:systematicity_qual_analysis} presents examples where both ViT-B-16 and ViT-L-14 retrieve the correct caption successfully in the SC split and where VT-B-16 fails in the UC split. Through this analysis, we find that the SC split for LAION-400M trained models is dominated by simple two-atom examples such as ``purple couch''. The UC split, however, contains more complex examples that involve relationships such as ``curtains on the window''. In particular, we find that the ViT-B-16 model struggles with the relationship ``on'' and often retrieves a wrong caption where ``on'' is replaced with ``off'' or where the object is replaced with an atomic foil. For example, ViT-B-16 retrieves ``plants on bob and plants off building'' incorrectly when the groundtruth caption is ``plants on a building''. Nevertheless, the rank of the groundtruth caption is often still within the top three. This explains the narrower gap in ViT-B-16' Recall@3 between Seen Compounds and Unseen Compounds. On the other hand, we see that ViT-L-14 continues to retrieve the correct caption even on the more challenging Unseen Compounds split, suggesting that a larger model size could improve compositional systematicity. 

\section{Additional Related Work}
\noindent\textbf{Evaluating learned representations} By analyzing the properties of pretrained representations, our work continues a tradition of research in Computer Vision~\cite{li-etal-2020-bert-vision, rosch-libovicky-2022-probing, milewski-etal-2022-finding, https://doi.org/10.48550/arxiv.2206.07835, https://doi.org/10.48550/arxiv.2204.03162, frank-etal-2021-vision} and Natural Language Processing~\cite{rogers-etal-2020-primer, jawahar-etal-2019-bert, tenney-etal-2019-bert, webson-pavlick-2022-prompt, hessel-schofield-2021-effective, pham-etal-2021-order} that probes characteristics of representations themselves rather than their performance on downstream tasks. 
 Instead of learning probes, we use retrieval for zero-shot evaluation in order to avoid scenarios where the learned probe compensates for the characteristics deficient in the original representations~\cite{hewitt-liang-2019-designing,belinkov-2022-probing,cao2021low,lepori2020picking,zhou-srikumar-2021-directprobe}.

%% file: tables/productivity_error_modes.tex
\begin{table*}[]
\centering
\caption{A list of the potential failure modes a vision-language model may encounter when parsing increasingly complex scenes, and the corresponding hard negatives generated in our test datasets.}
\resizebox{\linewidth}{!}{
\begin{tabular}{l l p{0.25\textwidth} p{0.35\textwidth} p{0.3\textwidth}}
    Dataset & Label & Error Mode  & Hard Negative & Example \\ \hline \hline
    Sys & \textsc{HN-Atom}   & Ignoring incorrect atoms.
    & \textbf{Atomic foils}. Replace a single atom with a mutually exclusive or antonymic atom, enforced by WordNet.              & \begin{tabular}[c]{@{}l@{}} A grill on top of the porch. \\ $\to$:  A grill \hnerror{underneath} the porch. \end{tabular}  \\ \hline
    Sys & \textsc{HN-Comp}  & Ignoring proper binding of atoms into compounds.
    & \textbf{Compound foils}. Split the correct atoms of a single compound over two compounds; fill in the partial compounds with atomic foils (see above). 
    \vspace{.1ex}
    & \begin{tabular}[c]{@{}l@{}} A pink car. \\ $\to$:  A \hnerror{blue} car and a pink \hnerror{toy}. \\ $\to$: A pink \hnerror{flower} and a \hnerror{black} car. \end{tabular}  \\ \hline \hline
    Prod & \textsc{HN-Atom}   & Ignoring incorrect atoms.
    & \textbf{Atomic foils}. Replace a single atom with a mutually exclusive or antonymic atom, enforced by WordNet.            & \begin{tabular}[c]{@{}l@{}}Yellow vase on top of television. \\ $\to$:  \hnerror{Red} vase on top of television. \\ $\to$: Yellow vase \hnerror{underneath} television. \\ $\to$:  Yellow vase on top of \hnerror{shelf}.\end{tabular} \\ \hline
    Prod & \textsc{HN-Swap}  & Ignoring proper binding of atoms.
    & \textbf{Swapping foils}. Swap two atoms of the same type -- or permute several atoms of the same type.
    & \begin{tabular}[t]{@{}l@{}}Yellow vase on top of television. \\ $\to$:  Yellow \hnerror{television} on top of \hnerror{vase}. \\ 
    $\to$:  Television on top of \hnerror{yellow} vase. \end{tabular} \\ \hline
    Prod & \textsc{HN-Neg}  & Disregarding incorrect negations.
    & \textbf{Negation foils}. Negate the entire caption \textit{or} an individual atom with a grammatically correct ``not" modifier. & \begin{tabular}[c]{@{}l@{}}Yellow vase on top of television. \\ $\to$:  \hnerror{There is no} yellow vase on top \\ of television. \\
    $\to$: Vase \hnerror{that is not yellow} on top \\ of television. \end{tabular}
\end{tabular}
    }
\label{tab:hard_negatives}
\end{table*}

%% file: tables/scene_graph_parser.tex
\begin{table*}[ht]
\small
\renewcommand{\arraystretch}{1.1}
\centering
\caption{\textit{Scene Graph Parser Validation:} We report precision and recall values the Stanford Scene Graph parser obtains on the CC-12M, YFCC-15M and LAION-400M datasets. For each dataset, we compute values for object, attribute and relationship atoms as well as object-relationship-object triplets. Overall, the scene graph obtains high precision values but lower recall scores. The parser performs the poorest on LAION-400M due its noisier captions.}
\begin{tabular}{lllllll}
             & \multicolumn{2}{l}{CC-12M} & \multicolumn{2}{l}{YFCC-15M} & \multicolumn{2}{l}{LAION-400M} \\ \cline{2-7} 
             & Precision     & Recall     & Precision      & Recall      & Precision       & Recall       \\ \hline \hline
Object       & 88.14         & 83.06      & 96.24          & 93.33       & 69.91           & 60.68        \\
Attribute    & 93.00            & 56.51      & 94.44          & 75.56       & 72.22           & 36.11        \\
Relationship & 92.86         & 70.18      & 93.59          & 83.33       & 88.33           & 40.15        \\
Triplet      & 91.67         & 64.04      & 92.31          & 81.11       & 87.00              & 39.55        \\ \hline
\end{tabular}
\label{tab:scene_graph_parser}
\end{table*}

%% file: tables/productivity_gt_annotation_breakdown.tex
\begin{table}[thb]
    \centering
    \caption{Productivity ground truth captions' faithfulness to their paired images, split by caption complexity. Overall, the generated captions' faithfulness is stable and consistently high across different complexities.}
    \begin{tabular}{c c}
        \textbf{Complexity} & \textbf{Avg faithfulness } \\ \hline \hline
        $n=7$\;\; & $88.7 \pm 10.8$\\
        $n=8$\;\; & $85.7 \pm 7.0$\\
        $n=9$\;\; & $90.0 \pm 6.0$ \\
        $n=10$ & $87.7 \pm 9.3$\\
        $n=11$ &  $88.1 \pm 7.8$ \\
        $n=12$ & $89.1 \pm 2.9$\\
    \end{tabular}    
    \label{tab:manual_eval_gt_complexity}
\end{table}

%% file: tables/image-text-counts.tex
\begin{table*}[ht]
\small
\renewcommand{\arraystretch}{1.1}
\centering
\caption{We report the ground truth caption counts in ${\D_{test}^{raw}}$ and hard negative counts in $\D_{test}^{HN}$s for systematicity and productivity, separated by hard negative type and split.}
\begin{tabular}{l r r r p{0.001\textwidth} l r r r r}
Systematicity & & & & & Productivity \\
\cmidrule{1-4}\cmidrule{6-10} 
Split & Ground Truth & \textsc{HN-Atom} & \textsc{HN-Comp} & & Split & Ground Truth & \textsc{HN-Atom} & \textsc{HN-Swap} & \textsc{HN-Neg} \\
\cmidrule{1-4}\cmidrule{6-10}
CC-12M SC & 262,541 & 104,024 & 156,036 & & $n = 4$ & 1,508
& 6,290 & 135 & 2,510 \\
CC-12M UC & 113,659 & 14,348 & 21,522 & & $n = 5$ & 1,734
& 7,270 & 180 & 3,425 \\
CC-12M UA & 9,577 & - & - & & $n = 6$ & 1,905
& 9,025 & 1,310 & 6,565 \\
YFCC SC & 194,502 & 75,948 & 113,922 & & $n = 7$ & 2,171
& 10,410 & 2,525 & 7,845 \\
YFCC UC & 172,469 & 39,204 & 58,806 & & $n = 8$ & 2,247
& 11,205 & 4,955 & 10,210 \\
YFCC UA & 18,806 & - & - & & $n = 9$ & 1,969
& 9,485 & 4,420 & 8,310 \\
LAION SC & 170,253 & 62,884 & 94,326 & & $n = 10$ & 2,246
& 11,325 & 6,465 & 10,460 \\
LAION UC & 201,595 & 49,604 & 74,406 & & $n = 11$ & 1,895
& 8,620 & 5,380 & 7,925 \\
LAION UA & 1,855 & - & - & & $n = 12$ & 1,878
& 10,005 & 7,890 & 9,710 \\
\end{tabular}
\label{tab:image_text_counts}
\end{table*}

%% file: tables/retrieval_set_size.tex
\begin{table}[]
\centering
\caption{ We summarize the retrieval set sizes for both $\D_{test}^{HN}$ and $\D_{test}^{Raw}$ in our systematicity and productivity evaluation.}
\resizebox{\columnwidth}{!}{
\begin{tabular}{ccccc}
Retrieval set size             & \multicolumn{3}{c}{$\D_{test}^{HN}$}             & $\D_{test}^{RAW}$ \\\hline \hline
\multirow{2}{*}{Systematicity} & \textsc{HN-Atom} & \textsc{HN-Comp} & & ---    \\
                               & 5       & 7 &       & 1,855   \\\hline
\multirow{2}{*}{Productivity}  &\textsc{HN-Atom} & \textsc{HN-Swap}       & \textsc{HN-Neg}     & ---    \\
                               & 6       & 6             & 6           & 1,508  
\end{tabular}
}
\label{tab:retrieval_set_size}
\end{table}

%% file: figures/syst-dataset-viz.tex
\begin{figure*}[ht!]
     \centering
     \includegraphics[width=\linewidth]{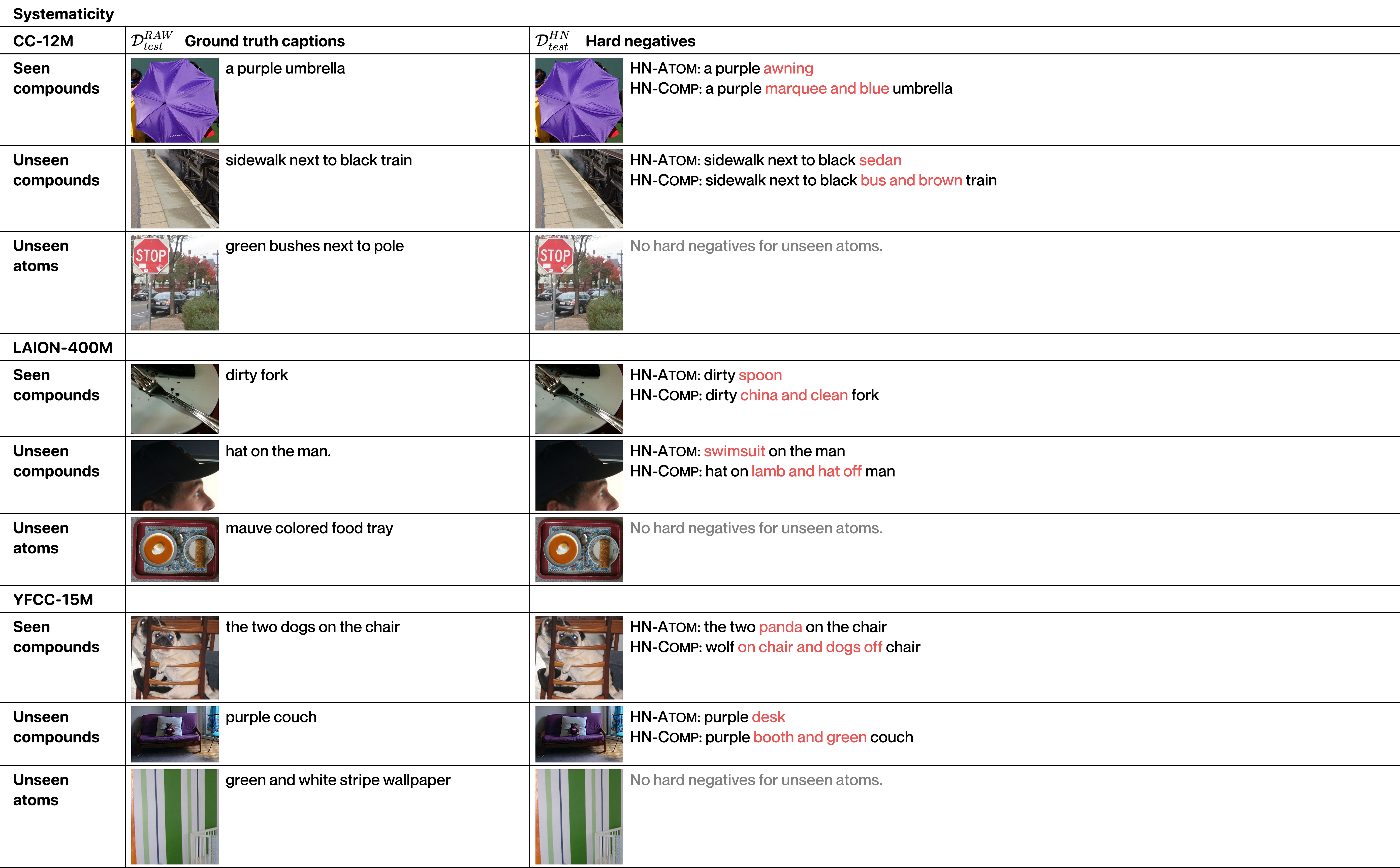}
     \caption{A sample of image-caption pairs in the \textbf{systematicity} retrieval sets. One ground truth caption is shown for each split of each training dataset, each of which lie in both $\D_{test}^{RAW}$ and $\D_{test}^{HN}$. Additionally, one example of each hard negative type is shown for each ground truth caption.}
     \label{fig:syst_viz}
 \end{figure*}

%% file: figures/prod-dataset-viz.tex
\begin{figure*}[ht!]
     \centering
     \includegraphics[width=\linewidth]{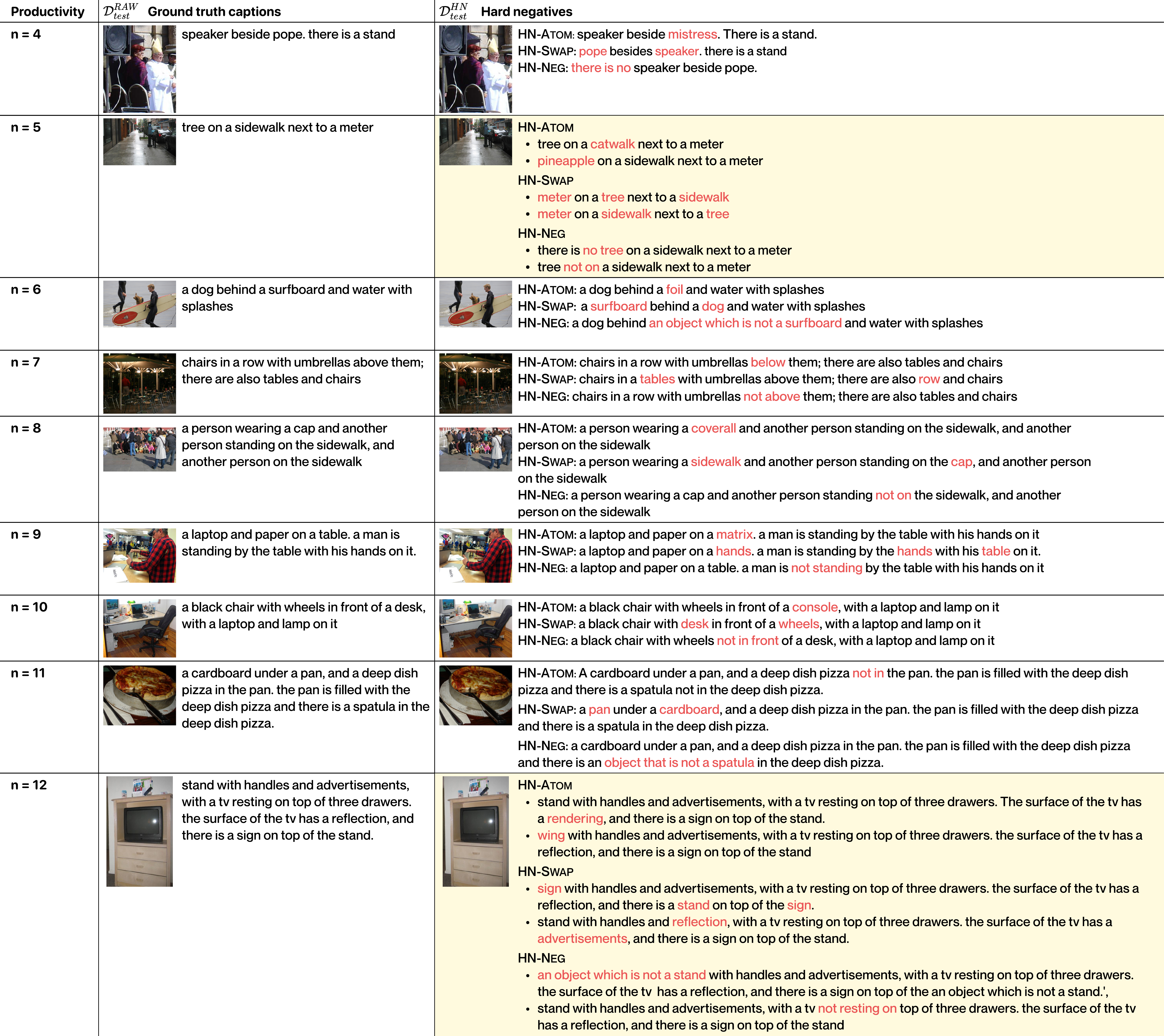}
     \caption{A sample of image-caption pairs in the \textbf{productivity} retrieval sets. One ground truth (GT) caption is shown for each complexity $n$. These GT captions lie in both $\D_{test}^{RAW}$ and $\D_{test}^{HN}$. One example of each hard negative type is shown for each GT caption. For two highlighted example captions ($n = 5, 12$), we show 2 hard negatives per type for comprehensiveness.}    
     \label{fig:prod_viz}
 \end{figure*}

%% file: tables/productivity_hard_negs_annotation_breakdown.tex
\begin{table}[thb]
    \centering
    \caption{Accuracy of our generated hard negatives for productivity, split by type, in our data verification. While \textsc{HN-Atom} atoms receive strong human evaluation scores, we find that \textsc{HN-Swap} and \textsc{HN-Neg} negatives are noisier.}
    \begin{tabular}{lcc}
        Type & Acc. mean $\pm$ std & Pairwise agreement\\ \hline \hline
        \textsc{HN-Atom} & $91.6 \pm 4.2$ & 83.1\\
        \textsc{HN-Swap} & $70.1 \pm 9.1$ & 58.5\\
        \textsc{HN-Neg} & $72.4 \pm 0.0$ & 59.5 \\
    \end{tabular}    
    \label{tab:manual_eval_prod_hard_negs}
\end{table}

%% file: tables/atom_comp_counts.tex
\begin{table}[]
\caption{ We summarize the unique atom and compound counts in the SC and UC split of the systematicity hard negative set.}
\resizebox{\columnwidth}{!}{
\begin{tabular}{lllll}
              & \multicolumn{2}{l}{SC}            & \multicolumn{2}{l}{UC}              \\
Train dataset & Atom (seen) & Comp (seen) & Atom (seen) & Comp (unseen) \\\hline
CC12M         & 3,348            & 26,006           & 946             & 3,587              \\
YFCC          & 3,173            & 18,987           & 1,405            & 9,801              \\
LAION         & 2,968            & 12,401           & 1,951            & 15,721            
\end{tabular}
}
\label{tab:atom_comp_counts}
\end{table}

%% file: figures/syst_atom_frequency.tex
\begin{figure*}
    \includegraphics[width=.45\linewidth]{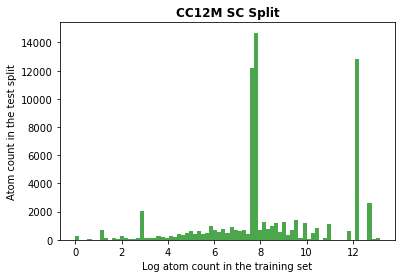}\hfill
    \includegraphics[width=.45\linewidth]{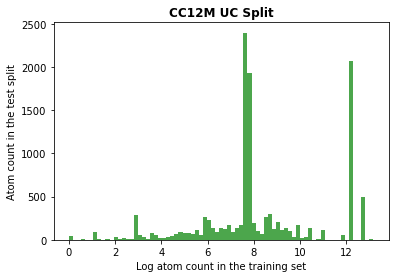}
    \\[\smallskipamount]

    \includegraphics[width=.45\linewidth]{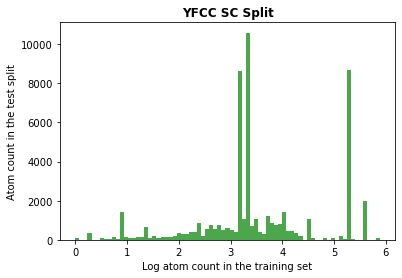}\hfill
    \includegraphics[width=.45\linewidth]{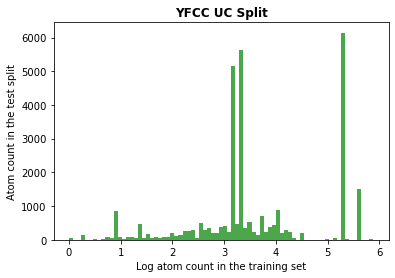}
    \\[\smallskipamount]

    \includegraphics[width=.45\linewidth]{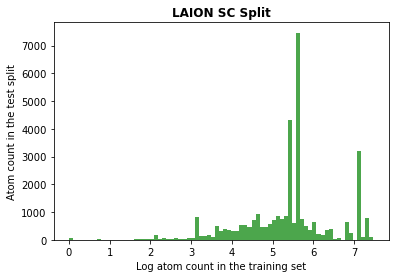}\hfill
    \includegraphics[width=.45\linewidth]{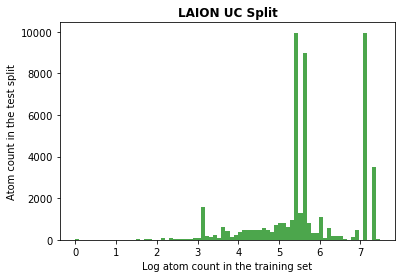}
    \caption{We plot the atom count in training vs. in the systematicity hard negative test set. We observe that the atoms in the SC and UC test splits have similar counts in the training dataset. }\label{fig:syst_atom_frequency}
\end{figure*}

%% file: tables/systematicity_hard_negs.tex
\begin{table*}[ht]
\small
\renewcommand{\arraystretch}{1.1}
\centering
\caption{\textit{Systematicity \textsc{HN-Atom} + \textsc{HN-Comp} Dataset Analysis.} We report Recall@1,3,5 and Avg R@K results for all models on the $D_{test}^{HN}$ hard-negative datasets with both \textsc{HN-Atom} + \textsc{HN-Comp}. Model performance decreases from the Seen all compounds (SC) to the Unseen Compounds (UC) split, particularly for LAION-400M models.}
\begin{tabular}{lllllllll}
                              & \multirow{2}{*}{Training dataset} & \multirow{2}{*}{Model} & \multicolumn{2}{c}{R@1} & \multicolumn{2}{c}{R@3} & \multicolumn{2}{c}{Avg R@K} \\
                              &                                   &                        & SC         & UC         & SC         & UC         & SC           & UC           \\\hline\hline
                              & & Random            & 9.09   & 9.09     & 27.27  & 27.27    & 18.18   & 18.18    \\\hline
\multirow{7}{*}{Image-to-text} & CC12M            & RN50              & 23.26       & 19.96       & 62.44       & 59.52       & 42.85           & 39.74      \\\cline{2-9}
 & \multirow{2}{*}{YFCC15M}          & RN50              & 23.38       & 20.08       & 60.09       & 56.61       & 41.74           & 38.34   \\
         &        & RN101             & 22.74       & 20.50       & 59.15       & 57.59       & 40.94           & 39.04    \\\cline{2-9}
& \multirow{4}{*}{LAION400M}        & ViT-B-32          & 34.28       & 28.00       & 70.74       & 68.74       & 52.51           & 48.37    \\
         &        & ViT-B-16          & 37.01       & 30.81       & 73.92       & 72.85       & 55.46           & 51.83     \\
          &       & ViT-B-16+240 & 37.32       & 32.26       & 75.03       & 73.46       & 56.17           & 52.86    \\
          &       & ViT-L-14          & 39.44       & 33.81       & 74.31       & 73.47       & 56.87           & 53.64   \\\hline

\end{tabular}
\label{tab:systematicity_hard_negs}
\end{table*}

%% file: tables/systematicity_hard_negs_atom.tex
\begin{table*}[ht]
\small
\renewcommand{\arraystretch}{1.1}
\centering
\caption{\textit{Systematicity \textsc{HN-Atom} Dataset Analysis.} We report Recall@1,3 and Avg R@K results for all models on the $D_{test}^{HN}$ subset with \textsc{HN-Atom}. Model performance decreases from the Seen Compounds (SC) to the Unseen Compounds (UC) split, particularly for LAION-400M models.}
\begin{tabular}{lllllllll}
                              & \multirow{2}{*}{Training dataset} & \multirow{2}{*}{Model} & \multicolumn{2}{c}{R@1} & \multicolumn{2}{c}{R@3} & \multicolumn{2}{c}{Avg R@K} \\
                              &                                   &                        & SC         & UC         & SC         & UC         & SC           & UC           \\\hline\hline
                              &                                   & Random       & 20.00 & 20.00 & 60.00 & 60.00 & 40.00   & 40.00         \\\hline
\multirow{7}{*}{Image-to-text} & CC12M                             & RN50                   & 39.26       & 34.88       & 88.81       & 88.10       & 64.04           & 61.49        \\\cline{2-9}
                              & \multirow{2}{*}{YFCC15M}          & RN50         & 43.35       & 39.50       & 90.55       & 90.07       & 66.95           & 64.78 \\
                              &                                   & RN101        & 43.26       & 39.85       & 90.33       & 90.30       & 66.79           & 65.08 \\\cline{2-9}
                              & \multirow{4}{*}{LAION400M}        & ViT-B-32     & 55.32       & 42.75       & 93.38       & 91.92       & 74.35           & 67.34 \\
                              &                                   & ViT-B-16     & 57.18       & 44.93       & 94.01       & 92.95       & 75.59           & 68.94 \\
                              &                                   & ViT-B-16+240 & 57.95       & 46.53       & 94.36       & 93.40       & 76.16           & 69.97  \\
                              &                                   & ViT-L-14     & 59.11       & 47.86       & 94.39       & 93.66       & 76.75           & 70.76 \\\hline 
\end{tabular}
\label{tab:systematicity_hard_negs_atom}
\end{table*}

%% file: tables/systematicity_hard_negs_comp.tex
\begin{table*}[ht]
\small
\renewcommand{\arraystretch}{1.1}
\centering
\caption{\textit{Systematicity \textsc{HN-COMP} Dataset Analysis.} We report Recall@1,3 and Avg R@K results for all models on the $D_{test}^{HN}$ subset with \textsc{HN-COMP}. We observe little to no difference in performance between the SC and UC split.}
\begin{tabular}{lllllllll}
                              & \multirow{2}{*}{Training dataset} & \multirow{2}{*}{Model} & \multicolumn{2}{c}{R@1} & \multicolumn{2}{c}{R@3} & \multicolumn{2}{c}{Avg R@K} \\
                              &                                   &                        & SC         & UC         & SC         & UC         & SC           & UC           \\\hline\hline
                                          &     & Random       & 14.29 & 14.29 & 42.86 & 42.86 & 28.57   & 28.57 \\\hline
\multirow{7}{*}{Image-to-text} & CC12M            & RN50         & 48.02       & 45.27       & 80.24       & 79.59       & 64.13           & 62.43  \\\cline{2-9}
                              & \multirow{2}{*}{YFCC15M}           & RN50         & 42.07       & 39.83       & 75.06       & 73.66       & 58.56           & 56.74 \\
             &    & RN101        & 40.72       & 39.56       & 74.71       & 74.16       & 57.72           & 56.86  \\\cline{2-9}
                              & \multirow{4}{*}{LAION400M}        & ViT-B-32     & 52.29       & 54.80       & 82.40       & 83.25       & 67.35           & 69.02  \\
               &  & ViT-B-16     & 56.00       & 59.00       & 84.64       & 86.24       & 70.32           & 72.62 \\
             &    & ViT-B-16+240 & 56.57       & 60.19       & 85.28       & 85.69       & 70.92           & 72.94 \\
                 & & ViT-L-14     & 57.10       & 60.78       & 84.17       & 85.69       & 70.64           & 73.24 \\\hline 
\end{tabular}
\label{tab:systematicity_hard_negs_comp}
\end{table*}

%% file: tables/productivity_hard_negs_atom.tex
\begin{table*}[ht]
\small
\renewcommand{\arraystretch}{1.1}
\centering
\caption{\textit{Productivity \textsc{HN-Atom} Dataset Analysis.} We report mean Recall@1 results for all models across all complexities. We find that models' Recall@1 values decrease as caption complexity increases.}
\begin{tabular}{lllrrrrrrrrr}
                               & Training dataset            & Model             & 4     & 5     & 6     & 7     & 8     & 9     & 10    & 11    & 12    \\\hline\hline
                               &                             & Random            & 16.67  & 16.67 & 16.67 & 16.67 & 16.67 & 16.67 & 16.67 & 16.67 & 16.67   \\\hline
\multirow{14}{*}{Image-to-text} & CC-12M                      & RN50              & 19.71  & 21.46 & 16.40 & 18.35 & 15.31 & 14.92 & 13.38 & 15.26 & 12.04  \\ \cline{2-12}
                               & \multirow{2}{*}{YFCC-15M}   & RN50              & 21.30  & 23.31 & 19.94 & 18.11 & 15.62 & 16.03 & 14.97 & 15.14 & 12.89 \\
                               &                             & RN101             & 22.66  & 22.21 & 18.17 & 18.44 & 15.22 & 16.34 & 15.85 & 17.34 & 13.04\\\cline{2-12}
                               & \multirow{4}{*}{LAION-400M} & ViT-B-32          & 23.21  & 21.25 & 19.06 & 18.59 & 16.15 & 13.92 & 15.76 & 15.49 & 12.84 \\
                               &                             & ViT-B-16 & 23.13  & 22.83 & 19.89 & 21.09 & 18.65 & 15.92 & 16.42 & 17.00 & 13.39 \\
                               &                             & ViT-B-16+240         & 29.73  & 23.31 & 21.72 & 21.81 & 18.38 & 18.08 & 17.62 & 18.39 & 14.89\\
                               &                             & ViT-L-14          & 28.54  & 25.10 & 21.55 & 24.06 & 19.81 & 18.61 & 17.88 & 18.68 & 15.44 \\\cline{2-12}
                               &                     & CyCLIP RN50              & 18.20  & 15.13 & 15.24 & 14.46 & 11.91 & 11.12 & 11.70 & 11.95 & 8.35\\ \cline{3-12}
                               &                       & FLAVA              & 29.17  & 16.23 & 14.13 & 15.08 & 14.46 & 14.55 & 14.88 & 15.72 & 14.79 \\ \cline{3-12}
                               &                       & ALBEF              & 38.71  & 32.94 & 27.87 & 27.76 & 26.51 & 25.94 & 25.92 & 27.03 & 24.34\\ \cline{2-12}
                               &     \multirow{4}{*}{CLIP's dataset}        & RN50              & 26.79  & 26.41 & 21.83 & 21.09 & 18.38 & 19.40 & 17.40 & 19.14 & 15.49\\
                               &                             & RN101             & 28.46  & 26.34 & 22.22 & 22.33 & 18.56 & 19.14 & 19.16 & 18.56 & 17.94\\
                               &  & ViT-B-32          & 28.70  & 23.31 & 21.33 & 19.79 & 18.61 & 18.13 & 17.66 & 18.10 & 16.69\\
                               &                             & ViT-B-16 & 30.68  & 26.41 & 23.93 & 23.15 & 19.19 & 19.19 & 18.76 & 20.19 & 16.44  \\
                               &                             & ViT-L-14          & 31.00  & 28.47 & 22.71 & 22.48 & 19.01 & 21.09 & 18.01 & 19.14 & 18.24 \\\hline
\end{tabular}
\label{tab:productivity_hard_negs_atom}
\end{table*}

%% file: tables/productivity_hard_negs_swap.tex
\begin{table*}[ht]
\small
\renewcommand{\arraystretch}{1.1}
\centering
\caption{\textit{Productivity \textsc{HN-Swap} Dataset Analysis.} We report mean Recall@1 results for all models across all complexities. We find that models' Recall@1 values are near or even below random chance across all complexities.}
\begin{tabular}{lllrrrrrrrrr}
                               & Training dataset            & Model             & 4     & 5     & 6     & 7     & 8     & 9     & 10    & 11    & 12    \\\hline\hline
                               &                             & Random            & 16.67  & 16.67 & 16.67 & 16.67 & 16.67 & 16.67 & 16.67 & 16.67 & 16.67   \\\hline
\multirow{14}{*}{Image-to-text} & CC-12M                      & RN50              & 7.41   & 19.44 & 9.92  & 12.87 & 12.61 & 13.12 & 13.53 & 14.87 & 12.93  \\ \cline{2-12}
                               & \multirow{2}{*}{YFCC-15M}   & RN50              & 11.11  & 16.67 & 14.50 & 15.25 & 13.32 & 14.14 & 10.75 & 13.48 & 13.88 \\
                               &                             & RN101             & 25.93  & 30.56 & 8.40  & 12.67 & 11.81 & 11.76 & 12.30 & 13.85 & 13.69 \\\cline{2-12}
                               & \multirow{4}{*}{LAION-400M} & ViT-B-32          & 25.93  & 19.44 & 12.98 & 13.86 & 12.31 & 14.37 & 11.60 & 12.08 & 14.13 \\
                               &                             & ViT-B-16 & 14.81  & 22.22 & 10.31 & 14.85 & 12.31 & 14.03 & 13.77 & 14.87 & 13.50 \\
                               &                             & ViT-B-16+240         & 22.22  & 22.22 & 14.12 & 15.05 & 14.13 & 14.48 & 13.69 & 18.31 & 15.91 \\
                               &                             & ViT-L-14          & 14.81  & 22.22 & 12.60 & 14.06 & 12.92 & 15.61 & 15.00 & 17.84 & 16.79\\\cline{2-12}
                               &                     & CyCLIP RN50              & 11.11 & 5.56  & 11.07 & 14.46 & 11.81 & 12.56 & 13.23 & 13.75 & 11.79\\ \cline{3-12}
                               &                       & FLAVA              & 7.41  & 19.44 & 9.16  & 10.50 & 9.69  & 11.65 & 10.44 & 12.36 & 16.22 \\ \cline{3-12}
                               &                       & ALBEF              & 25.93 & 13.89 & 17.56 & 20.00 & 21.19 & 19.80 & 20.42 & 22.12 & 22.43\\ \cline{2-12}
                               &     \multirow{4}{*}{CLIP's dataset}        & RN50              & 22.22 & 19.44 & 19.47 & 20.20 & 17.66 & 17.99 & 17.71 & 18.49 & 18.12 \\
                               &                             & RN101             & 29.63 & 25.00 & 16.79 & 17.62 & 17.15 & 15.38 & 17.40 & 19.61 & 18.06 \\
                               &  & ViT-B-32          & 25.93 & 22.22 & 22.90 & 15.64 & 15.14 & 16.63 & 16.24 & 20.91 & 18.95 \\
                               &                             & ViT-B-16 & 33.33 & 22.22 & 20.99 & 18.42 & 17.86 & 16.40 & 15.78 & 19.42 & 16.79 \\
                               &                             & ViT-L-14          & 11.11 & 13.89 & 19.08 & 16.83 & 17.05 & 16.86 & 16.01 & 18.49 & 18.06 \\\hline
\end{tabular}
\label{tab:productivity_hard_negs_swap}
\end{table*}

%% file: tables/productivity_hard_negs_negate.tex
\begin{table*}[ht]
\small
\renewcommand{\arraystretch}{1.1}
\centering
\caption{\textit{Productivity \textsc{HN-Neg} Dataset Analysis.} We report mean Recall@1 results for all models across all complexities. We find that models' Recall@1 values either stay near random chance or decrease as caption complexity increases except for some of OpenAI's CLIP models.}
\begin{tabular}{lllrrrrrrrrr}
                               & Training dataset            & Model             & 4     & 5     & 6     & 7     & 8     & 9     & 10    & 11    & 12    \\\hline\hline
                               &                             & Random            & 16.67  & 16.67 & 16.67 & 16.67 & 16.67 & 16.67 & 16.67 & 16.67 & 16.67   \\\hline
\multirow{14}{*}{Image-to-text} & CC-12M                      & RN50              & 15.13  & 19.28 & 18.41 & 23.59 & 20.94 & 20.83 & 21.95 & 20.90 & 18.34  \\ \cline{2-12}
                               & \multirow{2}{*}{YFCC-15M}   & RN50              & 8.32   & 10.29 & 12.18 & 12.64 & 12.30 & 12.23 & 12.65 & 13.49 & 12.96  \\
                               &                             & RN101             & 8.56   & 9.30  & 10.96 & 10.35 & 11.41 & 10.36 & 9.88  & 11.08 & 10.25\\\cline{2-12}
                               & \multirow{4}{*}{LAION-400M} & ViT-B-32          & 23.53  & 18.75 & 18.08 & 21.01 & 20.50 & 21.88 & 21.32 & 20.78 & 21.36\\
                               &                             & ViT-B-16 & 24.80  & 26.91 & 23.47 & 23.89 & 25.67 & 25.88 & 24.59 & 25.54 & 26.98 \\
                               &                             & ViT-B-16+240         & 23.53  & 28.20 & 26.36 & 29.12 & 28.83 & 27.03 & 28.30 & 30.30 & 27.49\\
                               &                             & ViT-L-14          & 30.35  & 29.04 & 26.14 & 29.92 & 28.48 & 27.41 & 28.70 & 30.72 & 29.60\\\cline{2-12}
                               &                     & CyCLIP RN50              & 16.24  & 12.65 & 15.35 & 14.39 & 13.19 & 12.77 & 13.23 & 12.77 & 13.27\\ \cline{3-12}
                               &                       & FLAVA              & 16.16  & 13.72 & 16.52 & 12.29 & 11.54 & 14.42 & 20.25 & 16.81 & 12.86\\ \cline{3-12}
                               &                       & ALBEF              & 75.83  & 45.12 & 43.55 & 44.05 & 47.28 & 48.19 & 46.45 & 47.71 & 40.05\\ \cline{2-12}
                               &     \multirow{4}{*}{CLIP's dataset}        & RN50              & 14.10  & 34.83 & 37.10 & 41.76 & 41.18 & 40.90 & 39.61 & 40.06 & 31.41\\
                               &                             & RN101             & 8.72   & 12.88 & 15.57 & 14.73 & 15.91 & 18.20 & 17.34 & 20.66 & 25.83\\
                               &  & ViT-B-32          & 15.85  & 30.56 & 34.09 & 35.99 & 38.99 & 40.13 & 39.87 & 41.33 & 38.94\\
                               &                             & ViT-B-16 & 7.13   & 26.22 & 28.64 & 32.11 & 33.16 & 31.41 & 34.29 & 36.51 & 31.56  \\
                               &                             & ViT-L-14          & 13.79  & 26.91 & 26.36 & 23.05 & 22.37 & 24.07 & 24.05 & 27.17 & 24.92 \\\hline
\end{tabular}
\label{tab:productivity_hard_negs_negate}
\end{table*}

%% file: tables/systematicity_weak_negs.tex
\begin{table*}[ht]
\small
\renewcommand{\arraystretch}{1.1}
\centering
\caption{\textit{Systematicity Raw Dataset Analysis.} We report mean Recall@1 results for all models across k-fold evaluations. Model performance consistently decreases from Seen all Compounds (SC) to Unseen Compounds (UC) and from Unseen Compounds to Unseen Atoms (UA) splits, particularly for LAION-400M models.}
\begin{tabular}{lllrrr}
                          
                            & Training dataset           & Model             & \multicolumn{1}{c}{SC} & \multicolumn{1}{c}{UC} & \multicolumn{1}{c}{UA} \\
\hline\hline
                            &                            & Random            & $0.05   \pm0.00$                             & $0.05   \pm0.00$                              & $0.05   \pm0.00$                            \\ \hline
\multirow{7}{*}{Image-to-text} & CC-12M                      & RN50              &$ 19.92\pm0.94$&$ 17.82\pm0.99$&$ 15.02\pm0.85$                \\ \cline{2-6}
                            & \multirow{2}{*}{YFCC-15M}   & RN50              &$ 16.30\pm0.70$&$ 14.57\pm0.69$&$ 12.80\pm0.90$                            \\
                            &                            & RN101             &$ 17.10\pm0.90$&$ 15.58\pm1.04$&$ 13.62\pm0.84$                            \\\cline{2-6}
                            & \multirow{4}{*}{LAION-400M} & ViT-B-16          &$ 35.61\pm0.92$&$ 30.04\pm1.42$&$ 25.88\pm0.00$                          \\
                            &                            & ViT-B-16+240 &$ 36.80\pm0.90$&$ 31.10\pm1.37$&$ 26.25\pm0.00$                            \\
                            &                            & ViT-B-32          &$ 33.86\pm0.97$&$ 29.00\pm1.40$&$ 23.99\pm0.00$                            \\
                            &                            & ViT-L-14          &$ 38.24\pm0.70$&$ 32.70\pm1.30$&$ 26.42\pm0.00$                           \\ \hline
\multirow{7}{*}{Text-to-image} & CC-12M                      & RN50              &$ 20.85\pm0.98$&$ 18.15\pm0.84$&$ 15.46\pm1.10$                           \\\cline{2-6}
                            & \multirow{2}{*}{YFCC-15M}   & RN50             &$ 15.60\pm0.79$&$ 14.05\pm0.84$&$ 12.17\pm0.64$                          \\
                            &                            & RN101            &$ 16.11\pm0.84$&$ 14.47\pm0.87$&$ 12.54\pm0.66$                            \\\cline{2-6}
                            & \multirow{4}{*}{LAION-400M} & ViT-B-16          &$ 35.74\pm0.76$&$ 29.58\pm1.39$&$ 23.29\pm0.00$                            \\
                            &                            & ViT-B-16+240& $ 37.25\pm0.97$&$ 30.57\pm1.33$&$ 24.26\pm0.00$                            \\
                            &                            & ViT-B-32          &$ 33.66\pm1.03$&$ 29.00\pm1.40$&$ 22.10\pm0.00$                            \\
                            &                            & ViT-L-14          &$ 38.69\pm0.86$&$ 32.00\pm0.90$&$ 25.61\pm0.00$ \\\hline                           
\end{tabular}
\label{tab:systematicity_weak_negs}
\end{table*}

%% file: tables/productivity_weak_negs.tex
\begin{table*}[ht]
\small
\renewcommand{\arraystretch}{1.1}
\centering
\caption{\textit{Productivity Raw Dataset Analysis.} We report mean Recall@1 results for all models across all complexities. We find that models' Recall@1 increases as the caption complexity increases.}
\begin{tabular}{llllllllllll}
                               & Training dataset            & Model             & 4     & 5     & 6     & 7     & 8     & 9     & 10    & 11    & 12    \\\hline\hline
                               &                             & Random            & 0.06  & 0.06  & 0.06  & 0.06  & 0.06  & 0.06  & 0.06  & 0.06  & 0.06  \\\hline
\multirow{7}{*}{Image-to-text} & CC-12M                      & RN50              & 13.19 & 14.23 & 16.20 & 19.02 & 21.17 & 19.26 & 22.70 & 24.29 & 25.77 \\ \cline{2-12}
                               & \multirow{2}{*}{YFCC-15M}   & RN50              & 9.08  & 10.92 & 12.09 & 14.23 & 14.17 & 15.40 & 14.54 & 16.75 & 19.02 \\
                               &                             & RN101             & 11.10 & 11.47 & 11.90 & 14.60 & 15.34 & 16.50 & 16.99 & 18.90 & 20.67 \\\cline{2-12}
                               & \multirow{4}{*}{LAION-400M} & ViT-B-16          & 20.80 & 20.00 & 25.89 & 27.30 & 29.63 & 29.02 & 30.31 & 34.11 & 37.36 \\
                               &                             & ViT-B-16+240 & 22.39 & 21.53 & 26.93 & 28.10 & 30.61 & 30.18 & 32.88 & 36.50 & 38.83 \\
                               &                             & ViT-B-32          & 20.49 & 20.37 & 23.50 & 26.93 & 29.20 & 28.22 & 29.94 & 32.58 & 35.40 \\
                               &                             & ViT-L-14          & 22.09 & 23.07 & 27.67 & 29.69 & 33.13 & 31.04 & 35.09 & 37.12 & 40.25 \\\hline
\multirow{7}{*}{Text-to-image} & CC-12M                      & RN50              & 12.52 & 15.03 & 15.52 & 17.85 & 17.30 & 19.75 & 21.66 & 22.52 & 25.58 \\\cline{2-12}
                               & \multirow{2}{*}{YFCC-15M}   & RN50              & 8.04  & 9.82  & 9.82  & 12.15 & 12.94 & 13.62 & 13.37 & 14.97 & 15.15 \\
                               &                             & RN101             & 9.39  & 11.10 & 11.10 & 13.31 & 13.74 & 15.03 & 14.48 & 16.07 & 18.40 \\\cline{2-12}
                               & \multirow{4}{*}{LAION-400M} & ViT-B-16          & 18.16 & 19.26 & 23.62 & 24.85 & 27.85 & 27.79 & 28.77 & 31.53 & 33.93 \\
                               &                             & ViT-B-16+240 & 18.96 & 20.67 & 25.46 & 26.13 & 28.83 & 29.02 & 31.41 & 32.88 & 37.24 \\
                               &                             & ViT-B-32          & 17.55 & 18.65 & 22.21 & 23.50 & 26.20 & 26.13 & 27.36 & 28.83 & 32.21 \\
                               &                             & ViT-L-14          & 19.88 & 20.43 & 24.97 & 26.63 & 30.37 & 30.00 & 32.76 & 34.42 & 36.99\\\hline
\end{tabular}
\label{tab:productivity_weak_negs}
\end{table*}

%% file: figures/productivity_hard_negs_overall.tex
\begin{figure}[t]
     \centering
     \includegraphics[width=0.8\columnwidth]{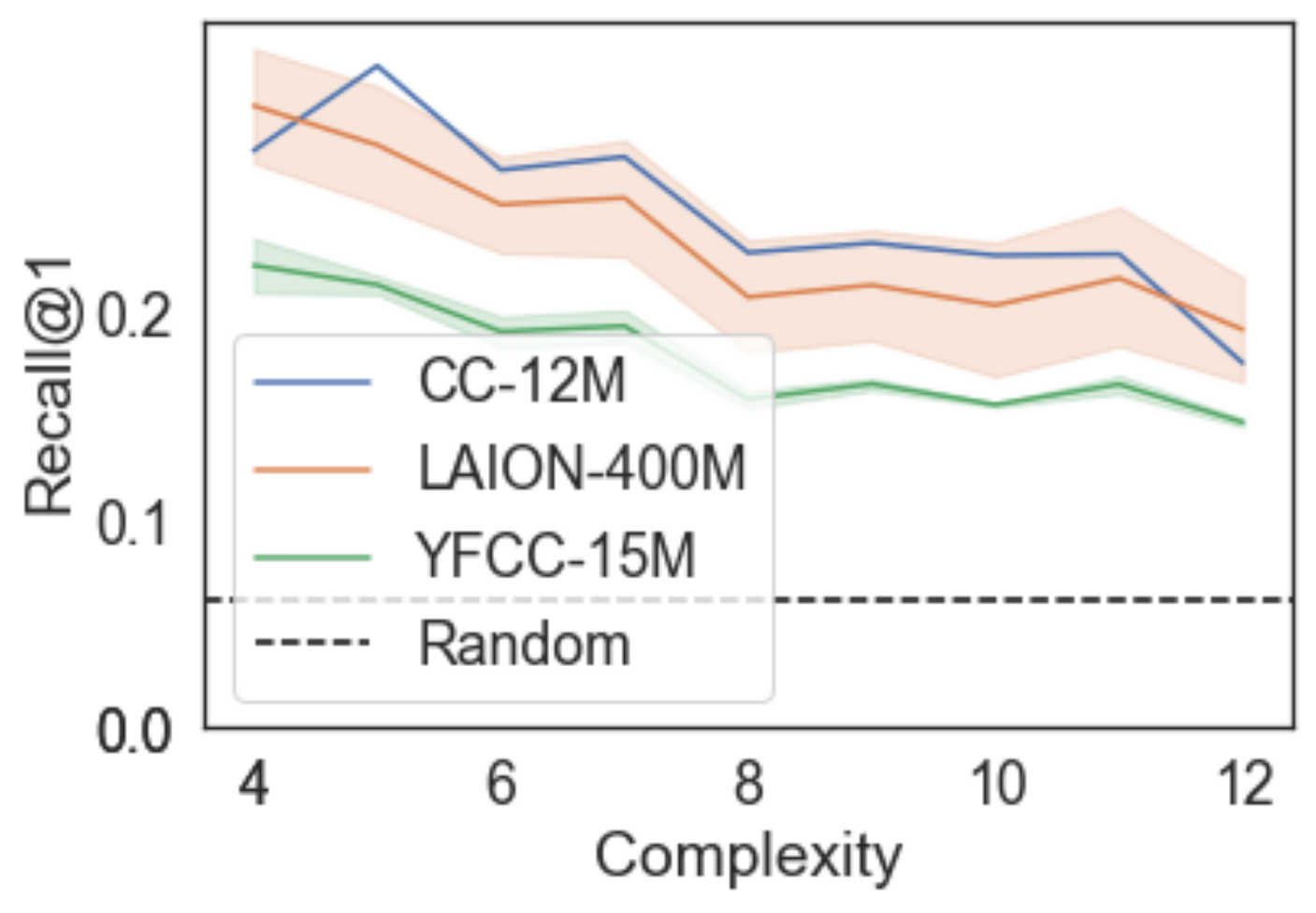}
     \caption{\textit{Productivity Analysis on Hard Negatives of All Types.} We plot models' Recall@1 on the overall hard negatives retrieval set against complexity, where each retrieval set contains hard negatives of all types. We find that models' ability to correctly retrieve the ground-truth caption drops as complexity increases.}
     \label{fig:prod_hard_negs_overall}
 \end{figure}
 

%% file: tables/systematicity_qualitative_analysis.tex
\begin{table*}[ht]
     \centering
     \caption{\textit{Systematicity Qualitative Analysis}. We present examples where LAION-400M trained ViT-B-16 and ViT-L-14 both perform well on the Seen Compounds (SC) split, and where ViT-B-16 performs poorly on the Unseen Compounds (UC) split.}
     \resizebox{\textwidth}{!}{
     \begin{tabular}{lllllll}
                                  &       &                                       & \multicolumn{2}{c}{ViT-B-16}                                                                                                                                                             & \multicolumn{2}{c}{ViT-L-14}                                                                                                                                                   \\
                                  & Image & GT caption                            & R@1 & Top 3 captions                                                                                                                                                              & R@1 & Top 3 captions                                                                                                                                                    \\\hline \hline
\multirow{5}{*}{SC}   &     \raisebox{-0.5\totalheight}{\includegraphics[width=0.2\textwidth]{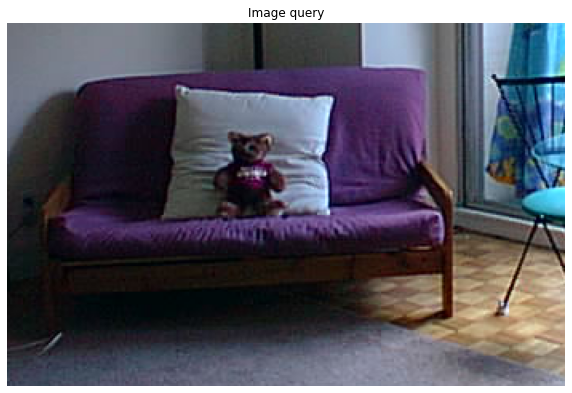}}  & purple couch                          & 1          & \begin{tabular}[c]{@{}l@{}}\correct{purple couch}\\ purple altar and brown couch\\ purple commode and red couch\end{tabular}                                                          & 1          & \begin{tabular}[c]{@{}l@{}} \correct{purple couch}\\ purple altar and brown couch\\ purple desk and brown couch\end{tabular}                                                 \\
                                  &    \raisebox{-0.5\totalheight}{\includegraphics[width=0.2\textwidth]{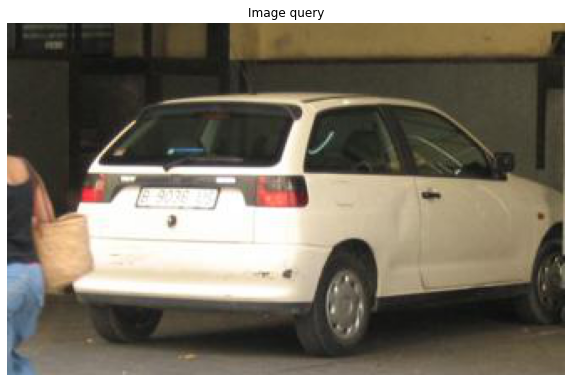}}   & a white parked car                 & 1          & \begin{tabular}[c]{@{}l@{}}\correct{a white parked car}\\ a green parked car\\ a white bike\end{tabular}                                                                              & 1          & \begin{tabular}[c]{@{}l@{}}\correct{a white parked car}\\ a green parked car\\ a orange parked car\end{tabular}                                                             \\
                                  &   \raisebox{-0.5\totalheight}{\includegraphics[width=0.2\textwidth]{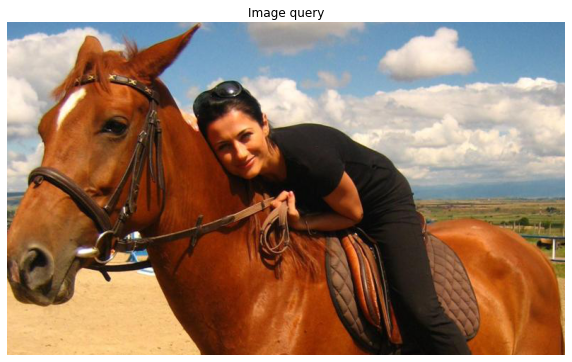}}    & a fully grown brown horse             & 1          & \begin{tabular}[c]{@{}l@{}}\correct{a fully grown brown horse}\\ a fully grown brown mule and red horse\\ a fully grown brown mule and yellow horse\end{tabular}                      & 1          & \begin{tabular}[c]{@{}l@{}}\correct{a fully grown brown horse}\\ a fully grown brown mule and red horse\\ a fully grown brown zebra and blue horse\end{tabular}             \\
                                  \\ \hline
\multirow{5}{*}{UC} &   \raisebox{-0.5\totalheight}{\includegraphics[width=0.2\textwidth]{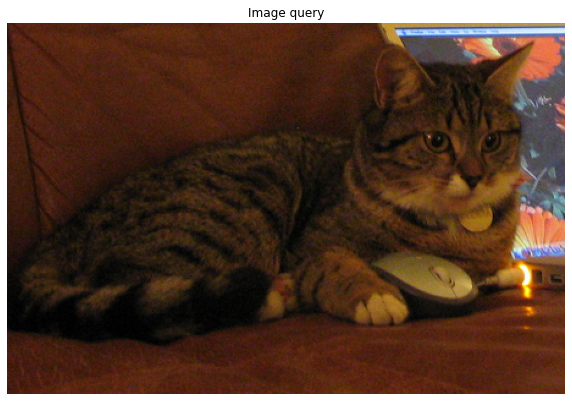}}    & a cat on the sofa.                    & 0          & \begin{tabular}[c]{@{}l@{}}a cat on the console.\\ \correct{a cat on the sofa}.\\ cat on counter and cat off sofa\end{tabular}                                                        & 1          & \begin{tabular}[c]{@{}l@{}}\correct{a cat on the sofa}.\\ a cat on the console.\\ badger on sofa and cat on console\end{tabular}                                            \\
                                  &    \raisebox{-0.5\totalheight}{\includegraphics[width=0.2\textwidth]{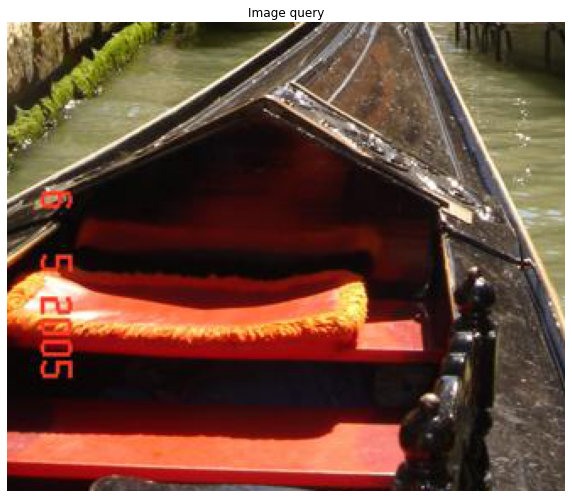}}   & boat on the water                     & 0          & \begin{tabular}[c]{@{}l@{}}boat on the polish\\ boat on the soda\\ \correct{boat on the water}\end{tabular}                                                                           & 1          & \begin{tabular}[c]{@{}l@{}}\correct{boat on the water}\\ boat on the lime\\ ship on water and boat on rubber\end{tabular}                                                   \\
                                  &    \raisebox{-0.5\totalheight}{\includegraphics[width=0.2\textwidth]{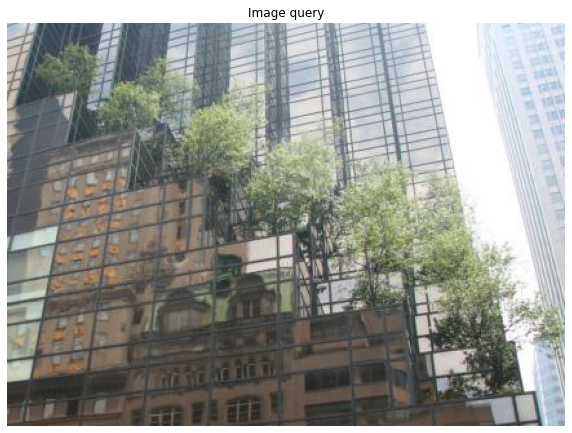}}   & plants on a building                  & 0          & \begin{tabular}[c]{@{}l@{}}plants on bob and plants off building\\ \correct{plants on a building}\\ court on building and plants off building\end{tabular}                            & 1          & \begin{tabular}[c]{@{}l@{}}\correct{plants on a building}\\ park on a building\\ billboard on building and plants off building\end{tabular}                                 \\

\end{tabular}}
      
      \label{tab:systematicity_qual_analysis}
      \end{table*}
  